\pdfoutput=1

\documentclass[11pt]{article}

\usepackage[]{acl}

\usepackage{times}
\usepackage{latexsym}
\usepackage{amsthm,amsmath,amssymb}
\usepackage{mathrsfs}
\usepackage{booktabs} 
\usepackage{multirow}
\usepackage[T1]{fontenc}

\usepackage[utf8]{inputenc}

\usepackage{microtype}

\usepackage{inconsolata}

\usepackage{graphicx}
\usepackage{enumitem}
\usepackage{subcaption} 
\usepackage{array}
\usepackage{makecell} 
\usepackage{tabularx} 

\usepackage{multirow}
\usepackage{colortbl}
\usepackage[normalem]{ulem}

\title{Dynamic Attention-Guided Context Decoding for Mitigating Context Faithfulness Hallucinations in Large Language Models}

\author{
  \textbf{Yanwen Huang}\textsuperscript{1,2,\textdagger}, 
  \textbf{Yong Zhang}\textsuperscript{1,\textdagger}, 
  \textbf{Ning Cheng}\textsuperscript{1,\textasteriskcentered},
  \\
  \textbf{Zhitao Li}\textsuperscript{1}, 
  \textbf{Shaojun Wang}\textsuperscript{1}, 
  \textbf{Jing Xiao}\textsuperscript{1}, 
  \\
  \textsuperscript{1} Ping An Technology (Shenzhen) Co., Ltd., China\textsuperscript{\textdagger}\\
  \textsuperscript{2} University of Electronic Science and Technology of China\textsuperscript{\textdagger}\\
    \texttt{\{zhangyong203, chengning211\}@pingan.com.cn}
}

\begin{document}
\maketitle

\renewcommand{\thefootnote}{}

\footnotetext{\textdagger\ Equal contribution.}
\footnotetext{* Corresponding author.}
\footnotetext{This work was done during Yanwen Huang’s internship at Ping An Technology (Shenzhen) Co., Ltd., China.}

\renewcommand{\thefootnote}{\arabic{footnote}}

\begin{abstract}

Large language models (LLMs) often exhibit Context Faithfulness Hallucinations, where outputs deviate from retrieved information due to incomplete context integration. Our analysis reveals a strong correlation between token-level uncertainty and hallucinations. We hypothesize that attention mechanisms inherently encode context utilization signals, supported by probing analysis. Based on these insights, we propose \textbf{Dynamic Attention-Guided Context Decoding (DAGCD)}, a lightweight framework that leverages attention distributions and uncertainty signals in a single-pass decoding. Experiments on open-book QA datasets demonstrate DAGCD’s effectiveness, yielding significant improvements in faithfulness and robustness while preserving computational efficiency.\footnote{Our code is available at \href{https://github.com/uestc-huangyw/DAGCD}{uestc-huangyw/DAGCD.}}

\end{abstract}

\section{Introduction}

Large Language Models (LLMs) \cite{brown2020gpt3, achiam2023gpt4, touvron2023llama2} excel in generating fluent and contextually relevant responses. However, they often struggle with factual accuracy, especially when relying on external information. \cite{static_knowledge}. Retrieval-Augmented Generation (RAG) \cite{rag1-2020, rag2-2020} mitigates this by grounding outputs in retrieved context, making it effective for tasks like question answering and reasoning \cite{rag-survey2023, rag-survey2024}. However, models often fail to faithfully utilize retrieved context, resulting in \textbf{Context Faithfulness Hallucinations}, where outputs deviate from the retrieved context \cite{hallucination-survey/faithfulness-hallucinate, hallucination-in-NLG}.

These hallucinations undermine the reliability of RAG systems, particularly in critical domains where factual accuracy is paramount\cite{chuang-etal-2024-lookbacklens}. Existing methods, such as CAD \cite{shi-etal-2024-CAD} and COIECD \cite{yuan-etal-2024-COIECD}, attempt to mitigate context faithfulness hallucinations by dynamically adjusting decoding distributions through token-level probability distribution comparisons or token-level uncertainty signals. While effective to some extent, these methods face several key limitations: limited interpretability, degraded performance when context-agnostic and context-aware outputs differ significantly, and computational complexity due to multiple decoding passes.

\begin{figure}
    \centering
    \includegraphics[width=\linewidth]{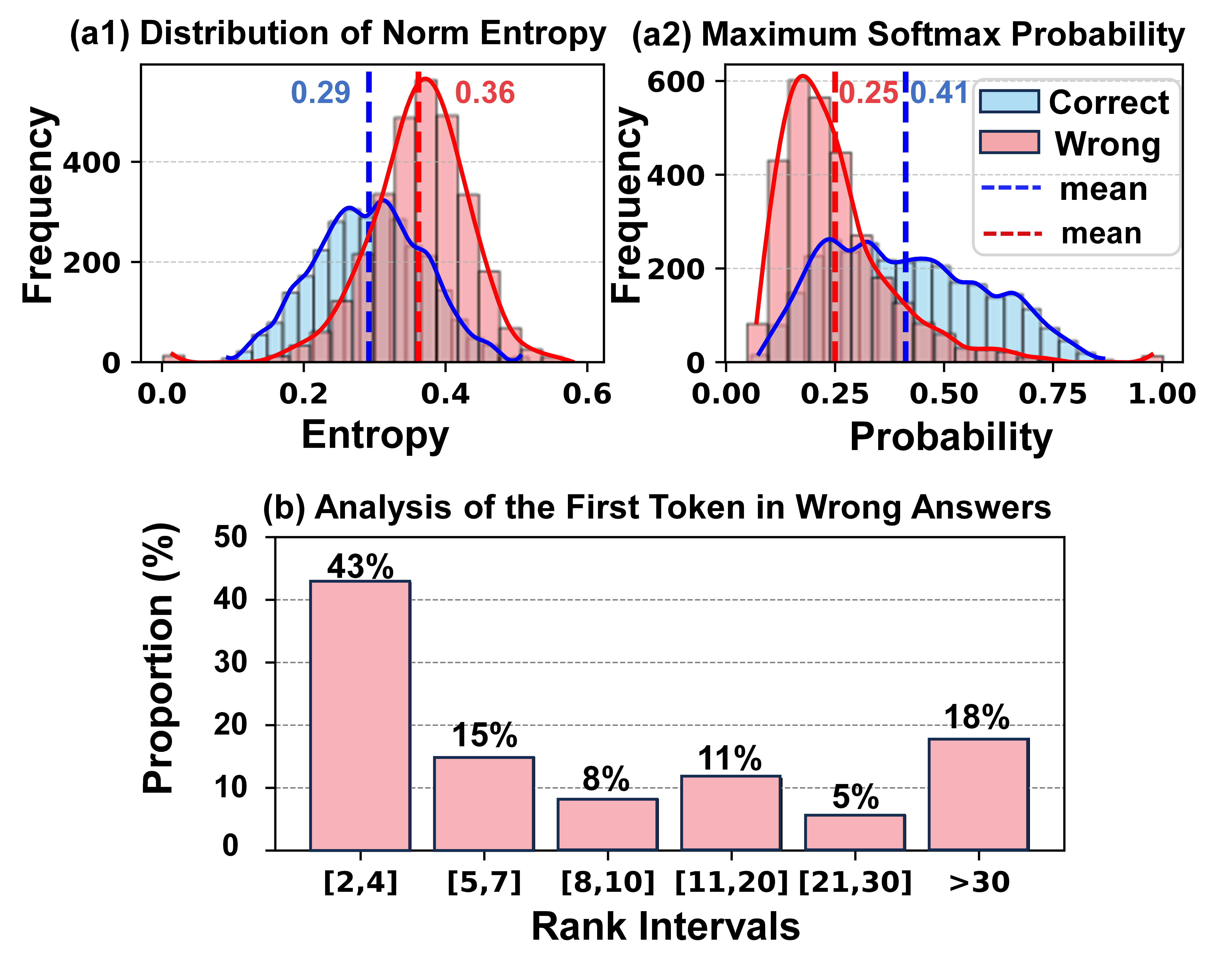}
    \vspace{-5mm}
    \caption{Analysis of the token-level probability distribution after context concatenation: (a1, a2) the model's uncertainty when generating correct versus wrong answers, measured by NE and MSP; (b) for wrong answers, the ranking of the golden answer token within the token-level probability distribution.}
    \label{fig: entropy and rank analysis}
    \vspace{-6mm}
\end{figure}

To better understand context faithfulness hallucinations and explore potential solutions, we take an internally-driven approach, analyzing intrinsic model signals that may explain why retrieval-augmented large language models struggle to utilize retrieved context faithfully \cite{liang2024internal}. Motivated by prior research linking token-level uncertainty to factual hallucinations~\cite{chuang-2024-Dola, das-etal-2025-entropy}, we examine whether token-level probability distribution entropy correlates with context faithfulness hallucinations.

Our analysis reveals a strong correlation between higher uncertainty and context faithfulness hallucinations. Specifically, as shown in Figure \ref{fig: entropy and rank analysis} (a1), wrong answers exhibit higher entropy in the token-level probability distribution, indicating greater uncertainty in generation. Even for correct answers (Figure \ref{fig: entropy and rank analysis} (a2)), it often assigns low confidence to the highest-ranked token, suggesting incomplete integration of retrieved context. Notably, in incorrect responses, most gold answer tokens appear in the top 10 in the token-level probability distribution, but are not assigned the highest probability (Figure \ref{fig: entropy and rank analysis} (b)), implying that \textbf{the model identifies relevant context but fails to prioritize it effectively}.

These findings indicate that while models retrieve relevant context, they struggle to integrate and prioritize it during generation. Since most gold answer tokens appear within the top 10, an effective strategy to mitigate context faithfulness hallucinations is to dynamically identify and prioritize these tokens during generation. This requires detecting reliable signals that indicate how retrieved context influences the model's predictions. Attention mechanisms in Transformer models naturally emerge as a key source of such signals, since they facilitate information flow between tokens \cite{induction_head, rome, 2023-attention-hypothesis-1}. We hypothesize that attention distributions encode intrinsic indicators of context utilization.

To validate our hypothesis, we trained a probing classifier using Logistic Regression on attention distributions, achieving over 0.99 AUC in distinguishing contextually relevant tokens. Even with just 100 training samples, the classifier demonstrated strong generalization across in-domain and cross-domain test sets, indicating that attention distributions inherently encode context utilization signals. These results support attention-based context utilization as a fundamental mechanism in LLMs. By leveraging these intrinsic signals, attention distributions provide a lightweight and interpretable means to assess how models integrate retrieved context into their predictions.

Motivated by these findings, we propose Dynamic Attention-Guided Context Decoding (DAGCD), a novel method to mitigate context faithfulness hallucinations. Inspired by the copy-generator framework~\cite{pointer-networks/CNN_DM,xu-etal-2020-copy-dist2}, DAGCD integrates attention weights to estimate the relevance of tokens in the retrieved context, dynamically adjusting output probabilities. Additionally, token-level uncertainty guides these adjustments by emphasizing underconfident yet contextually relevant tokens. By combining these strategies, DAGCD ensures output alignment with the retrieval context and maintains efficiency.

\noindent Our contributions are as follows:  
\vspace{-3mm}
\begin{enumerate}
    \item \textbf{Comprehensive analysis of context faithfulness hallucinations:} We identify a strong correlation between token-level uncertainty and context faithfulness hallucinations, showing that incorrect responses exhibit higher entropy and retrieved context is often recognized but not effectively prioritized.  \vspace{-3mm}

    \item \textbf{Attention-driven interpretability framework:} We propose \textbf{Dynamic Attention-Guided Context Decoding (DAGCD)}, leveraging attention distributions to amplify contextually relevant tokens and ensure faithful utilization of retrieved context. \vspace{-3mm}

    \item \textbf{Lightweight and efficient decoding:} DAGCD operates in a single decoding pass, integrating attention signals and uncertainty measures without additional overhead, improving efficiency. \vspace{-3mm}

    \item \textbf{Extensive validation across datasets and models:} DAGCD outperforms greedy decoding across multiple QA datasets, improving EM by \textbf{17.67\%} on pretrained models and \textbf{2.25\%} on instruction-tuned models, demonstrating robustness and scalability.

\end{enumerate}


\section{Why Can't Generate Faithful Answers?}
\label{sec2}

Token-level uncertainty is closely related to factual hallucinations, as models often exhibit higher entropy in the token-level probability distribution when generating factually incorrect outputs~\cite{chuang-2024-Dola, das-etal-2025-entropy}. While uncertainty measures help detect hallucination-prone predictions, most studies focus on factual hallucinations, where responses are incorrect without retrieved context. In contrast, context faithfulness hallucinations arise when models rely on retrieved information but generate misaligned or contradictory outputs. Despite their significance in RAG, their relationship with uncertainty remains unclear. Inspired by prior research, we investigate whether unfaithful responses in RAG exhibit higher entropy and whether contextually relevant tokens are recognized but assigned insufficient confidence.

\subsection{Uncertainty Leads to Unfaithful Answers} \label{sec:uncertainty}

\noindent \paragraph{Experimental Setup} To assess the relationship between uncertainty and response faithfulness, we use two common metrics: \textbf{Normalized Entropy (NE)}, which measures the overall uncertainty in the token-level probability distribution \cite{ entropy-uncertainty-2023}, and \textbf{Maximum Softmax Probability (MSP)}, which quantifies the model’s confidence in the highest probability within the token-level probability distribution \cite{maximum_softmax_probability(MSP)2017}. Higher NE indicates greater uncertainty, while higher MSP corresponds to greater confidence in the predictions of the model. For detailed experimental descriptions, see Appendix~\ref{app:uncertainty-details}.

\noindent \paragraph{Results and Analysis}  

Figure \ref{fig: entropy and rank analysis} (a1) and (a2) illustrates the Normalized Entropy and Maximum Softmax Probability of the token-level probability distribution when the model produces correct and wrong answers. The model exhibits higher uncertainty for wrong answers, with an average NE of 0.36 compared to 0.29 for correct cases, and a lower average MSP of 0.25 compared to 0.41. Notably, there is a substantial overlap between the correct (blue) and wrong (red) cases in the figure, indicating that even correct answers often exhibit high uncertainty. We also analyzed the correlation between prediction accuracy and uncertainty, and the results demonstrate a significant negative correlation, further confirming that \textbf{token-level uncertainty is strongly associated with unfaithful answers}. For detailed results see Appendix \ref{app: spearman}.

\subsection{LLM is Actually Utilizing Context} \label{sec:contextual_utilization}

Our previous analysis links token-level uncertainty to unfaithful responses, showing that incorrect outputs often have higher entropy and lower confidence. However, this does not mean the model ignores retrieved context. A key question remains: \textbf{Do incorrect responses imply that LLMs have failed to leverage the retrieved context?}

To investigate this, we examine the ranking of gold answer tokens in the token-level probability distribution for incorrect responses. If these tokens frequently rank high but are not assigned the highest probability, it suggests the model identifies relevant context but fails to prioritize it effectively.

\noindent \paragraph{Experimental Setup}

We analyze incorrect responses from §\ref{sec:uncertainty} by examining the ranking distribution of gold answer tokens in the token-level probability distribution of the first generated token after context concatenation. The ranks are grouped into intervals, and we compute the proportion of gold answer tokens within each rank interval.

\noindent \paragraph{Results and Analysis}  

As shown in Figure \ref{fig: entropy and rank analysis} (subplot b), when the model generates an incorrect response, 43\% of cases have the gold answer token ranked 2nd to 4th based on token probabilities, and 66\% have it ranked within the top 10 in the token-level probability distribution. Moreover, as shown in Figure \ref{app-fig-probs-gap}, the average probability gap between the gold answer token and the highest-probability token remains relatively small: 0.14 for ranks between 2 and 4, and 0.24 for ranks beyond 30.  

These findings indicate that \textbf{in context faithfulness hallucination scenarios, the model recognizes relevant context tokens but fails to prioritize them effectively, limiting their impact on the generated output}. This highlights an incomplete integration of retrieved context, emphasizing the need for improved context incorporation strategies to enhance faithfulness and mitigate hallucinations.


\section{Context Utilization Signal in Attention}
\label{sec:attention_signals}

Our analysis in Section \ref{sec2} highlights that while models often identify relevant context tokens, they fail to assign them sufficient confidence, leading to context faithfulness hallucinations. To better understand this phenomenon, we seek reliable signals that indicate which retrieved context tokens are effectively utilized by the model during generation.  

Due to their role in integrating and propagating information across tokens, attention mechanisms naturally emerge as a key candidate for capturing such signals \cite{induction_head, rome, 2023-attention-hypothesis-1}. We hypothesize that \textbf{attention distributions encode intrinsic indicators of context utilization}, providing a lightweight and interpretable means to assess how models incorporate retrieved context into their outputs.

\subsection{Attention Ratio}

A key challenge in analyzing attention weights is the noise from non-context tokens (e.g., delimiters) caused by attention sink effects \cite{attention-for-special-tokens-1, attention_sink}. Additionally, attention magnitudes vary across heads and layers, complicating feature comparison \cite{attention_ratio_1, attention_ratio_2}. To address these issues, we introduce the \textbf{attention ratio}, a normalized measure that captures how much attention a retrieved context token receives relative to the total attention assigned within the context. For a given token \(j\) in the retrieved context \(C\), the attention ratio at the $h$-th head in the $l$-th layer is defined as:
\begin{equation}
    r_{l,h}^j = \frac{a_{l,h}^j}{\sum_{j \in C} a_{l,h}^j}
\end{equation}
where \(a_{l,h}^j\) denotes the raw attention weight assigned to token \(j\). This ratio quantifies the relative importance of \(j\) within the retrieved context for a specific attention head.

To construct token-level features, we aggregate attention ratios across all heads:
\begin{equation}
    v_j = \left[ r_{1,1}^j, \dots, r_{\text{num\_layers},\text{num\_heads}}^j \right]
\end{equation}
This feature vector represents the distribution of attention across layers and heads, enabling a structured assessment of context token importance.

\begin{figure}[t]
    \centering
    \includegraphics[width=\linewidth]{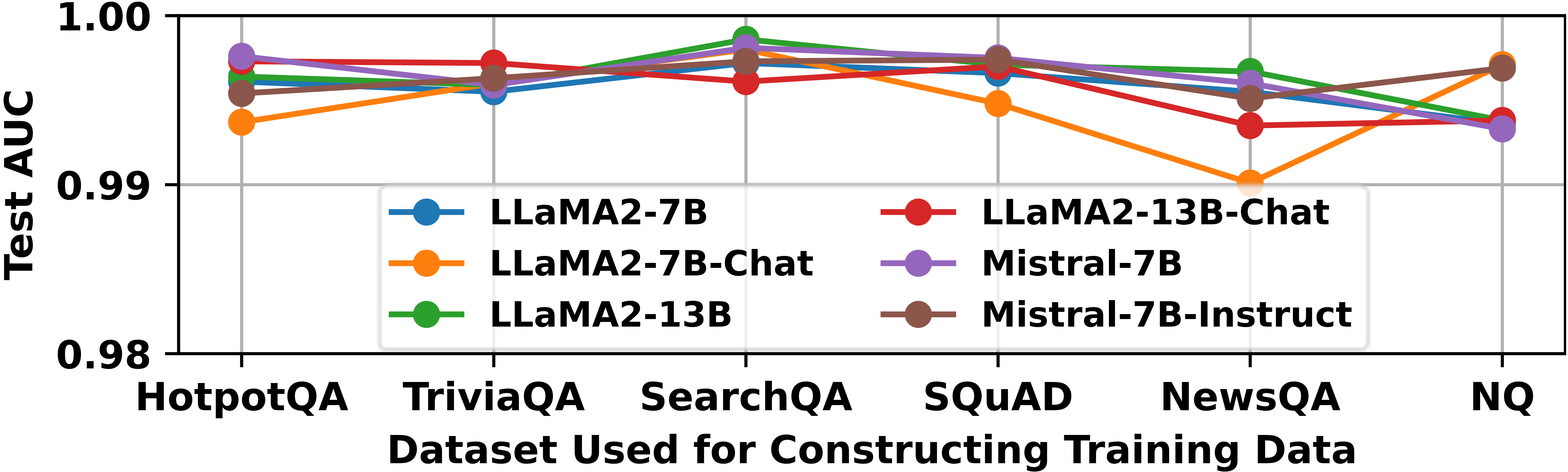}
    \caption{Cross Domain Validation. One dataset as the training set (X-axis) and the remaining datasets as the test sets, showing the AUC on the test sets (Y-axis).}
    \label{fig: out-domain LR}
    \vspace{-5mm}
\end{figure}

\subsection{Experimental Setup}
\label{sec:3 Experimental Setup}

To validate our hypothesis and investigate whether this mechanism generalizes across different datasets, training sizes, and prompt templates, we conduct the following experiments.

\noindent \textbf{Data Construction} We constructed the dataset by randomly selecting samples from the MrQA training set dataset (contains six open-book QA datasets in different domains) \cite{fisch-etal-2019-mrqa}, focusing on cases where the model’s output changed from incorrect to correct after context concatenation \cite{output_shifted_from_incorrect_2022}. These cases indicate that the model successfully leveraged the retrieved context to produce the correct answer. Context tokens were labeled as positive (utilized) if they corresponded to the gold answer, and negative (non-utilized) otherwise. Using these labels, we extracted attention ratio feature vectors $\mathbf{v}_j$ to train a Logistic Regression (LR) classifier.

\vspace{-2mm}
\subsection{Results}\label{sec:3 hypothesis-validation-results}
\vspace{-2mm}

\noindent \paragraph{Cross Domain Validation}
We tested the classifier on six sub-datasets from different domains. Specifically, we selected one dataset as the training set and tested the performance on the remaining datasets (each sub-dataset construct 500 samples, contains 250 positive and 250 negative samples).

As shown in Figure~\ref{fig: out-domain LR}, the classifier achieves an average AUC above 0.99 across all datasets and LLMs. The results indicate that the context utilization signal in attention is \textbf{data-independent}.

\noindent \paragraph{Training Data Size Validation}
Building on the cross domain experiment, we further tested the impact of training set size on classifier performance. Specifically, we trained the model using data constructed from one sub-dataset and evaluated its performance on the remaining sub-datasets. 

\begin{figure}[t]
    \centering
    \includegraphics[width=\linewidth]{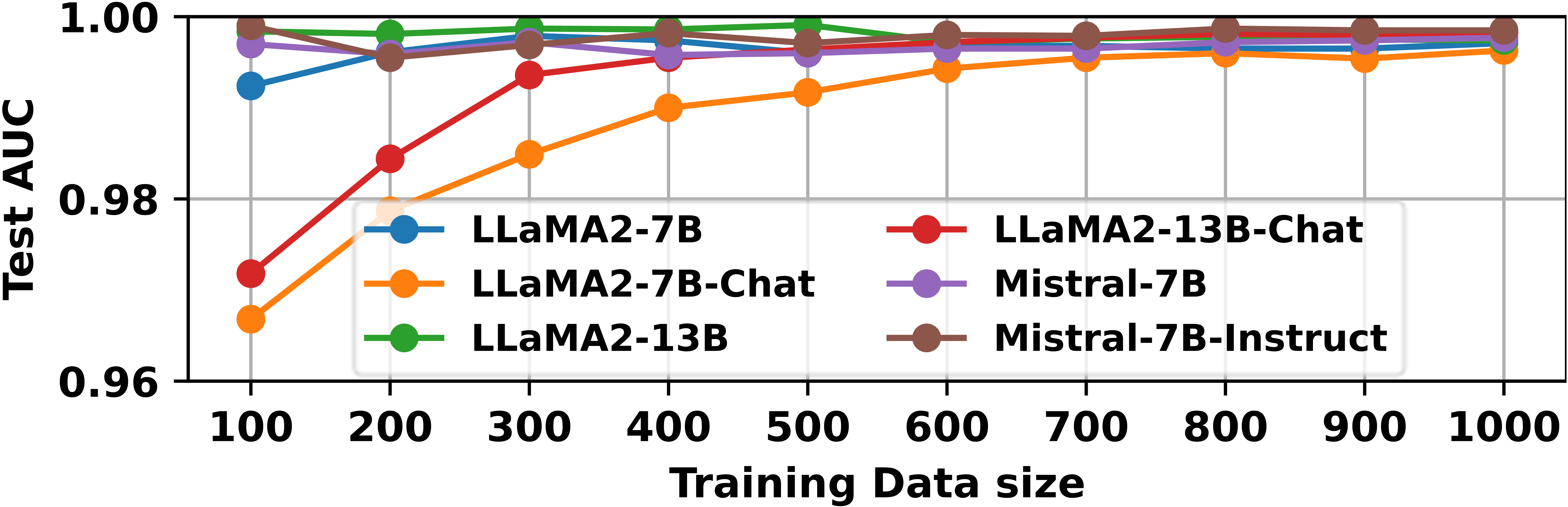}
    \caption{Training Data Size Validation. Training sets of varying sizes were constructed from a single dataset (HotpotQA), and evaluate on remain datasets.}
    \label{fig: LR-train-size-AUC}
    \vspace{-5mm}
\end{figure}

Figure \ref{fig: LR-train-size-AUC} shows the variation in model AUC with different training data sizes (where "Train size = 100" refers to a training set constructed with 50 positive and 50 negative samples). The results indicate that even with only 100 samples, the model's AUC exceeds 0.96, and the performance improvement becomes limited as the data size increases. This demonstrates that the LR classifier trained using the attention ratio exhibits strong \textbf{data-efficiency}.

\noindent \paragraph{Additional Results}

We then examined the impact of different prompt templates, and the results indicate that the classifier consistently maintains high performance regardless of the prompt template used. For Details, see Appendix \ref{app:section3-details}.

To better apply the classifier to practical tasks, we conducted a detailed analysis of the importance of different features and the relationships between them. The results show that \textbf{the classifier using the top-K features outperforms the one using the full feature set}. Furthermore, the attention heads exhibit Concentration and Complementarity features. For Details, see Appendix \ref{app: Characteristics of Different Features}.

\noindent \paragraph{Conclusion: A Fundamental Mechanism in Transformer-based LLMs}  
The consistent generalization of attention-based context utilization across datasets, data sizes, and prompts reinforces its role as a fundamental mechanism in LLMs. Our findings show that attention heads encode robust context integration signals, providing a lightweight and interpretable way to assess how models incorporate retrieved context.

\begin{figure*}[!htp]
    \centering
    \includegraphics[width=\linewidth]{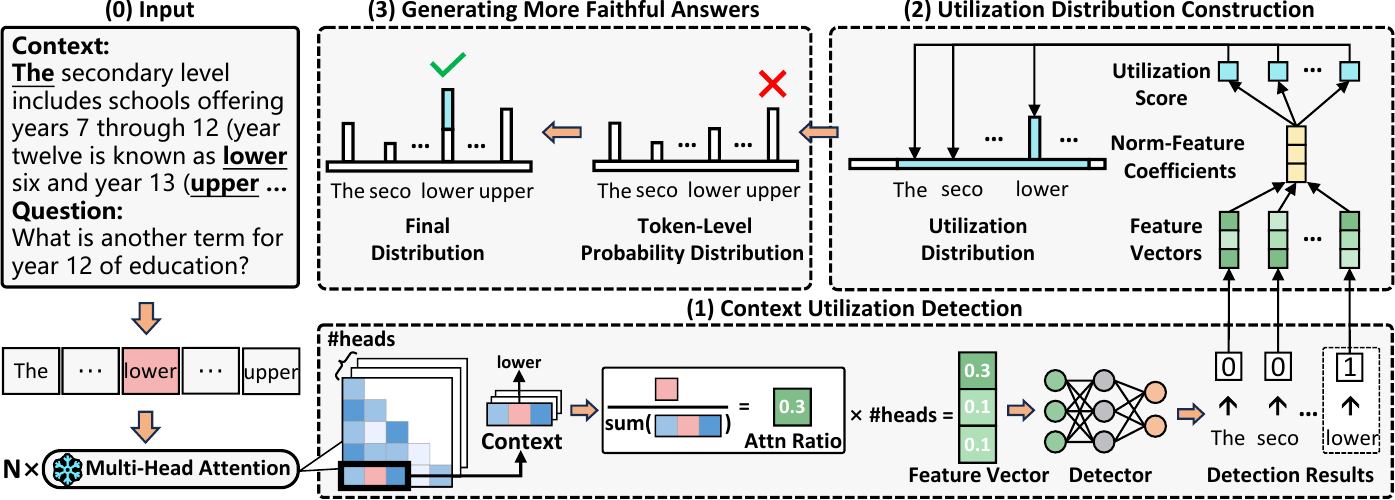}
    \caption{The illustration of the generation process of our proposed DAGCD method.}
    \label{fig:model}
    \vspace{-5mm}
\end{figure*}

\vspace{-2mm}
\section{Method}
\vspace{-2mm}

Inspired by the copy-generation mechanism~\cite{pointer-networks/CNN_DM,xu-etal-2020-copy-dist2}, we propose \textbf{Dynamic Attention-Guided Context Decoding (DAGCD)} to mitigate context faithfulness hallucinations. DAGCD leverages utilization signals to dynamically guide the generation process, focusing on relevant contextual tokens. It integrates three steps: detecting utilized context tokens during inference (§\ref{4.1}), constructing a utilization distribution (§\ref{4.2}), and adjusting the token-level probability distribution to enhance contextual utilization (§\ref{sec: 4.3 Generate More Faithful Answers}).

\subsection{Context Utilization Detection} 
\label{4.1}

\noindent \textbf{Context Utilization Detector}
To identify contextually relevant tokens during inference, we utilize a Logistic Regression (LR) classifier trained on attention-based utilization signals. Our analysis in Section \ref{sec:3 hypothesis-validation-results} shows that selecting the most informative attention heads improves generalization. Thus, we construct the Context Utilization Detector based on the top-K most important attention heads.

\noindent \textbf{Feature Data Collection}
To obtain feature vectors, we extract attention distributions at the current decoding step, as illustrated in Figure \ref{fig:model}. Specifically, we take the last row of the attention map for each selected head \( h_k \in H \) ($H$ is the set of top-K attention heads). To ensure focus on relevant information, we filter out non-contextual tokens, such as query tokens and placeholder tokens in templates.

The top-K feature vector for each contextual token \( j \) is then constructed as:
\begin{equation}
    \mathbf{v}_j^{(K)} = [r_{h_1}^j, r_{h_2}^j, \dots, r_{h_K}^j]
\end{equation}

where \( r_{h_k}^j \) represents the normalized attention ratio of token \( j \) in attention head \( h_k \). 

Finally, the feature vector $\mathbf{v}_j^{(K)}$ is fed into the detector, which identifies the set of context tokens actively utilized at the current decoding step.

\subsection{Utilization Distribution Construction} 
\label{4.2}

The context utilization detector identifies which tokens are utilized but does not quantify the degree of utilization for each token. To address this, we compute a utilization score \( s_j \) for each token \( j \) by aggregating attention ratios from selected attention heads, weighted by their normalized feature coefficients \( w_k \). Tokens classified as unused by the detector are directly assigned a score of zero.
\begin{equation}
    s_j = \sum_{k=1}^K (r_{h_k}^j \times w_k), \ w_k = \frac{c_k}{\sum_{k=1}^K c_k}
\end{equation}
where \( r_{h_k}^j \) is the normalized attention ratio of token \( j \) in attention head \( h_k \), and \( w_k \) is the importance weight assigned to head \( h_k \). The coefficient \( c_k \) is learned from the LR classifier, representing the contribution of each head to context utilization.

The utilization distribution \( U \) represents a probability distribution over context tokens, normalized based on their utilization scores:
\begin{equation}
    \mathbf{U} = [u_1, u_2, \dots, u_N], \quad u_i = \frac{s_i}{\sum_{j=1}^N s_j}
\end{equation}

where \( u_i \) denotes the utilization probability of token \( i \), \( N \) is the vocabulary size. Tokens either absent from the context or classified as non-utilized by the detector (\( s_i = 0 \)) are assigned \( u_i = 0 \).

\noindent \paragraph{Top-Rank Constraint}  
To enhance the reliability of generation adjustments, our approach applies a top-rank restriction, ensuring modifications focus on plausible tokens. Specifically, we define \( U_{\text{top}} \) as the subset of the utilization distribution \( U \) corresponding to tokens ranked within the top-\( R \) positions of the generation distribution.  

This design builds on prior work~\cite{li-etal-2023-CD, chuang-2024-Dola}, addressing context faithfulness hallucination challenges while leveraging our observation that correct context tokens usually appear within the top-ranked positions in the token-level probability distribution, even when the model generates incorrect answers. Through this constraint, we reduce the risk of amplifying irrelevant or nonsensical tokens, preserving output integrity.

\subsection{Generating More Faithful Answers}
\label{sec: 4.3 Generate More Faithful Answers}

DAGCD adjusts token probabilities based on token-level uncertainty, measured using the normalized entropy \( H_{\text{norm}}(P) \) of the token-level probability distribution. High entropy indicates greater uncertainty and a higher risk of generating contextually inconsistent responses. Since entropy correlates with uncertainty but lacks a fixed numerical relationship, we introduce a scaling factor \( \alpha \) to compensate for model-specific entropy variations.

Adjustments are applied only when utilized tokens in \( U_{\text{top}} \) overlap with the top-ranked tokens in the token-level probability distribution. The adjusted generation distribution \( P' \) is computed as:
\begin{equation}
    P' = P + \alpha H_{\text{norm}}(P) \cdot U_{\text{top}}
\end{equation}
where \( P \) represents the original token-level probability distribution, and \( \alpha H_{\text{norm}}(P) \) dynamically scales the adjustment based on model uncertainty.

\section{Experiments}

\begin{table*}[ht]
  \belowrulesep=0pt  
  \aboverulesep=0pt  
  \centering
  \resizebox{\linewidth}{!}{  
    \begin{tabular}{c|c|cc|cc|cc|cc|cc|cc|cc|cc}
    \toprule[1.5pt]
    \multirow{2}{*}{\textbf{Dataset}} & \multirow{2}{*}{\textbf{Decoding}} & \multicolumn{2}{c|}{\textbf{HotpotQA}} & \multicolumn{2}{c|}{\textbf{TriviaQA}} & \multicolumn{2}{c|}{\textbf{SearchQA}} & \multicolumn{2}{c|}{\textbf{SQuAD}} & \multicolumn{2}{c|}{\textbf{NewsQA}} & \multicolumn{2}{c|}{\textbf{NQ}} & \multicolumn{2}{c|}{\textbf{NQ-swap}} & \multicolumn{2}{c}{\textbf{Average}} \\
    \cmidrule{3-18}          &       & \textbf{EM} & \textbf{F1} & \textbf{EM} & \textbf{F1} & \textbf{EM} & \textbf{F1} & \textbf{EM} & \textbf{F1} & \textbf{EM} & \textbf{F1} & \textbf{EM} & \textbf{F1} & \textbf{EM} & \textbf{F1} & \textbf{EM} & \textbf{F1} \\
    \midrule  
    \multirow{4}{*}{\textbf{LLaMA2-7B}} & Greedy & \cellcolor[rgb]{ .949,  .949,  .949}\underline{44.74}  & \cellcolor[rgb]{ .949,  .949,  .949}\underline{54.71}  & \cellcolor[rgb]{ .949,  .949,  .949}55.28  & \cellcolor[rgb]{ .949,  .949,  .949}68.01  & \cellcolor[rgb]{ .949,  .949,  .949}54.21  & \cellcolor[rgb]{ .949,  .949,  .949}59.24  & \cellcolor[rgb]{ .949,  .949,  .949}39.90  & \cellcolor[rgb]{ .949,  .949,  .949}52.61  & \cellcolor[rgb]{ .949,  .949,  .949}32.93  & \cellcolor[rgb]{ .949,  .949,  .949}45.46  & \cellcolor[rgb]{ .949,  .949,  .949}\underline{38.85}  & \cellcolor[rgb]{ .949,  .949,  .949}50.59  & \cellcolor[rgb]{ .949,  .949,  .949}36.03  & \cellcolor[rgb]{ .949,  .949,  .949}36.62  & \cellcolor[rgb]{ .949,  .949,  .949}43.13  & \cellcolor[rgb]{ .949,  .949,  .949}52.46  \\
          & CAD   & 44.13  & 54.49  & 55.26  & 68.04  & 54.14  & 59.21  & 38.35  & 51.12  & 31.74  & 43.70  & 38.14  & 48.58  & \underline{36.11}  & \underline{36.69}  & 42.55  & 51.69  \\
          & COIECD & 42.03  & 51.48  & \underline{57.04}  & \underline{70.06}  & \textbf{57.03} & \textbf{63.14} & \underline{40.93}  & \underline{54.78}  & \underline{34.40}  & \underline{48.48}  & 38.79  & \underline{51.69}  & 34.98  & 35.63  & \underline{43.60}  & \underline{53.61}  \\
          & \textbf{OURs} &  \textbf{47.35} & \textbf{57.43} & \textbf{58.12} & \textbf{70.97} & \underline{54.35}  & \underline{59.70}  & \textbf{48.02} & \textbf{60.02} & \textbf{36.51} & \textbf{49.06} & \textbf{47.74} & \textbf{60.23} & \textbf{53.12} & \textbf{53.63} & \textbf{49.32} & \textbf{58.72} \\
    \midrule  
    \multirow{4}[0]{*}{\textbf{\makecell{LLaMA2-7B\\-Chat}}} & Greedy & \cellcolor[rgb]{ .949,  .949,  .949}\underline{53.33}  & \cellcolor[rgb]{ .949,  .949,  .949}\underline{67.41}  & \cellcolor[rgb]{ .949,  .949,  .949}\underline{71.72}  & \cellcolor[rgb]{ .949,  .949,  .949}\underline{76.83}  & \cellcolor[rgb]{ .949,  .949,  .949}54.19  & \cellcolor[rgb]{ .949,  .949,  .949}58.11  & \cellcolor[rgb]{ .949,  .949,  .949}67.69  & \cellcolor[rgb]{ .949,  .949,  .949}78.70  & \cellcolor[rgb]{ .949,  .949,  .949}39.41  & \cellcolor[rgb]{ .949,  .949,  .949}53.90  & \cellcolor[rgb]{ .949,  .949,  .949}50.47  & \cellcolor[rgb]{ .949,  .949,  .949}65.48  & \cellcolor[rgb]{ .949,  .949,  .949}67.98  & \cellcolor[rgb]{ .949,  .949,  .949}68.78  & \cellcolor[rgb]{ .949,  .949,  .949}57.83  & \cellcolor[rgb]{ .949,  .949,  .949}67.03  \\
          & CAD   & 52.86  & 67.16  & 71.70  & \underline{76.83}  & 54.16  & 58.11  & 65.89  & 77.91  & 38.46  & 53.26  & 48.89  & 65.00  & 68.04  & 68.85  & 57.14  & 66.73  \\
          & COIECD & 53.14  & 67.03  & \textbf{72.26} & \textbf{77.33} & \textbf{55.04} & \textbf{58.98} & \underline{68.32}  & \underline{79.56}  & \underline{40.00}  & \underline{54.55}  & \textbf{52.39} & \underline{66.84}  & \underline{69.48}  & \underline{70.13}  & \textbf{58.66} & \underline{67.77}  \\
          & \textbf{OURs} & \textbf{55.31} & \textbf{68.61} & 69.21  & 75.95  & \underline{54.25}  & \underline{58.13}  & \textbf{68.49} & \textbf{79.76} & \textbf{40.53} & \textbf{54.81} & \underline{51.69}  & \textbf{66.92} & \textbf{69.50} & \textbf{70.30} & \underline{58.43}  & \textbf{67.78} \\
    \midrule 
    \multirow{4}[0]{*}{\textbf{LLaMA2-13B}} & Greedy & \cellcolor[rgb]{ .949,  .949,  .949}\underline{52.36}  & \cellcolor[rgb]{ .949,  .949,  .949}\underline{63.40}  & \cellcolor[rgb]{ .949,  .949,  .949}58.25  & \cellcolor[rgb]{ .949,  .949,  .949}69.95  & \cellcolor[rgb]{ .949,  .949,  .949}\underline{63.22}  & \cellcolor[rgb]{ .949,  .949,  .949}\underline{68.33}  & \cellcolor[rgb]{ .949,  .949,  .949}51.64  & \cellcolor[rgb]{ .949,  .949,  .949}64.57  & \cellcolor[rgb]{ .949,  .949,  .949}30.84  & \cellcolor[rgb]{ .949,  .949,  .949}40.11  & \cellcolor[rgb]{ .949,  .949,  .949}42.26  & \cellcolor[rgb]{ .949,  .949,  .949}54.08  & \cellcolor[rgb]{ .949,  .949,  .949}49.02  & \cellcolor[rgb]{ .949,  .949,  .949}49.59  & \cellcolor[rgb]{ .949,  .949,  .949}49.66  & \cellcolor[rgb]{ .949,  .949,  .949}58.58  \\
          & CAD   & 51.53  & 63.11  & 58.25  & 69.92  & 63.13  & 68.32  & 49.94  & 63.44  & 29.94  & 39.04  & 41.40  & 52.27  & \underline{49.07}  & \underline{49.63}  & 49.04  & 57.96  \\
          & COIECD & 50.21  & 60.96  & \underline{59.19}  & \underline{71.49}  & \textbf{65.97} & \textbf{71.00} & \underline{52.73}  & \underline{65.93}  & \textbf{35.66} & \textbf{50.58} & \underline{42.35}  & \underline{54.37}  & 48.35  & 48.84  & \underline{50.64}  & \underline{60.45}  \\
          & \textbf{OURs} & \textbf{53.13} & \textbf{64.21} & \textbf{59.65} & \textbf{72.07} & 61.41  & 67.08  & \textbf{56.54} & \textbf{68.56} & \underline{33.26}  & \underline{42.20}  & \textbf{55.34} & \textbf{71.23} & \textbf{65.86} & \textbf{66.19} & \textbf{55.03} & \textbf{64.51} \\
    \midrule 
    \multirow{4}[0]{*}{\textbf{\makecell{LLaMA2-13B\\-Chat}}} & Greedy & \cellcolor[rgb]{ .949,  .949,  .949}55.01  & \cellcolor[rgb]{ .949,  .949,  .949}69.92  & \cellcolor[rgb]{ .949,  .949,  .949}\textbf{74.58} & \cellcolor[rgb]{ .949,  .949,  .949}\underline{79.35}  & \cellcolor[rgb]{ .949,  .949,  .949}67.08  & \cellcolor[rgb]{ .949,  .949,  .949}\underline{71.96}  & \cellcolor[rgb]{ .949,  .949,  .949}68.26  & \cellcolor[rgb]{ .949,  .949,  .949}79.45  & \cellcolor[rgb]{ .949,  .949,  .949}40.20  & \cellcolor[rgb]{ .949,  .949,  .949}55.11  & \cellcolor[rgb]{ .949,  .949,  .949}53.49  & \cellcolor[rgb]{ .949,  .949,  .949}69.18  & \cellcolor[rgb]{ .949,  .949,  .949}60.69  & \cellcolor[rgb]{ .949,  .949,  .949}61.77  & \cellcolor[rgb]{ .949,  .949,  .949}59.90  & \cellcolor[rgb]{ .949,  .949,  .949}69.53  \\
          & CAD   & 54.44  & 69.66  & \textbf{74.58} & \textbf{79.37} & 67.01  & 71.95  & 66.77  & 78.70  & 39.44  & 54.44  & 52.89  & 68.63  & 60.83  & 61.92  & 59.42  & 69.24  \\
          & COIECD & \underline{56.15}  & \underline{70.43}  & \underline{73.87}  & 78.96  & \underline{67.28}  & 71.93  & \underline{68.49}  & \underline{80.39}  & \underline{40.75}  & \underline{56.16}  & \underline{53.69}  & \underline{69.81}  & \underline{62.47}  & \underline{63.21}  & \underline{60.39}  & \underline{70.13}  \\
          & \textbf{OURs} & \textbf{57.76} & \textbf{71.69} & 73.04  & 78.77  & \textbf{68.19} & \textbf{72.73} & \textbf{69.66} & \textbf{80.76} & \textbf{40.78} & \textbf{56.24} & \textbf{55.36} & \textbf{71.31} & \textbf{64.03} & \textbf{65.20} & \textbf{61.26} & \textbf{70.96} \\
    \midrule 
    \multirow{4}[0]{*}{\textbf{Mistral-7B}} & Greedy & \cellcolor[rgb]{ .949,  .949,  .949}\underline{53.41}  & \cellcolor[rgb]{ .949,  .949,  .949}\underline{64.36}  & \cellcolor[rgb]{ .949,  .949,  .949}\textbf{59.45} & \cellcolor[rgb]{ .949,  .949,  .949}\underline{68.39}  & \cellcolor[rgb]{ .949,  .949,  .949}\underline{63.79}  & \cellcolor[rgb]{ .949,  .949,  .949}67.77  & \cellcolor[rgb]{ .949,  .949,  .949}44.19  & \cellcolor[rgb]{ .949,  .949,  .949}56.11  & \cellcolor[rgb]{ .949,  .949,  .949}31.51  & \cellcolor[rgb]{ .949,  .949,  .949}38.94  & \cellcolor[rgb]{ .949,  .949,  .949}33.74  & \cellcolor[rgb]{ .949,  .949,  .949}51.18  & \cellcolor[rgb]{ .949,  .949,  .949}39.80  & \cellcolor[rgb]{ .949,  .949,  .949}46.04  & \cellcolor[rgb]{ .949,  .949,  .949}46.56  & \cellcolor[rgb]{ .949,  .949,  .949}56.11  \\
          & CAD   & 41.57  & 56.01  & \underline{57.88}  & 67.48  & 63.64  & \underline{68.65}  & 34.08  & 47.55  & 25.78  & 35.37  & 23.18  & 41.46  & 26.96  & 35.97  & 39.01  & 50.36  \\
          & COIECD & 46.43  & 58.32  & 44.30  & 51.77  & 54.82  & 59.17  & \underline{50.50}  & \underline{60.98}  & \underline{40.05} & \underline{52.78}  & \underline{42.12}  & \underline{56.58}  & \underline{59.53}  & \underline{61.89}  & \underline{48.25}  & \underline{57.36}  \\
          & \textbf{OURs}  &    \textbf{63.45} & \textbf{73.89} & 56.76  & \textbf{71.86} & \textbf{64.49} & \textbf{69.88} & \textbf{63.04} & \textbf{73.75} & \textbf{41.62} & \textbf{55.13} & \textbf{57.85} & \textbf{71.19} & \textbf{69.46} & \textbf{69.86} & \textbf{59.52} & \textbf{69.37} \\
    \midrule 
    \multirow{4}[1]{*}{\textbf{\makecell{Mistral-7B\\-Instruct}}} & Greedy & \cellcolor[rgb]{ .949,  .949,  .949}58.70  & \cellcolor[rgb]{ .949,  .949,  .949}72.18  & \cellcolor[rgb]{ .949,  .949,  .949}69.64  & \cellcolor[rgb]{ .949,  .949,  .949}75.61  & \cellcolor[rgb]{ .949,  .949,  .949}44.42  & \cellcolor[rgb]{ .949,  .949,  .949}49.63  & \cellcolor[rgb]{ .949,  .949,  .949}67.28  & \cellcolor[rgb]{ .949,  .949,  .949}79.37  & \cellcolor[rgb]{ .949,  .949,  .949}39.79  & \cellcolor[rgb]{ .949,  .949,  .949}54.72  & \cellcolor[rgb]{ .949,  .949,  .949}52.29  & \cellcolor[rgb]{ .949,  .949,  .949}66.93  & \cellcolor[rgb]{ .949,  .949,  .949}66.90  & \cellcolor[rgb]{ .949,  .949,  .949}67.83  & \cellcolor[rgb]{ .949,  .949,  .949}\underline{57.00}  & \cellcolor[rgb]{ .949,  .949,  .949}66.61  \\
          & CAD   & 49.30  & 64.81  & \textbf{70.23} & \underline{75.95}  & \underline{45.42}  & \underline{50.96}  & 59.97  & 72.92  & 34.97  & 51.90  & 42.63  & 58.49  & 52.04  & 53.75  & 50.65  & 61.25  \\
          & COIECD & \underline{59.74}  & \underline{72.59}  & 64.92  & 72.15  & 37.09  & 42.66  & \textbf{68.45} & \textbf{81.03} & \underline{40.84} & \underline{55.96} & \underline{53.54}  & \underline{68.72}  & \textbf{72.81} & \textbf{73.81} & 56.77  & \underline{66.70}  \\
          & \textbf{OURs} & \textbf{60.55} & \textbf{73.49} & \underline{69.70}  & \textbf{75.96} & \textbf{47.17} & \textbf{52.65} & \underline{68.30}  & \underline{80.62}  & \textbf{40.85} & \textbf{56.23} & \textbf{54.78} & \textbf{69.72} & \underline{71.38}  & \underline{72.12}  & \textbf{58.96} & \textbf{68.68} \\
    \bottomrule[1.5pt]
    \end{tabular}
}
   \caption{Performance comparison of different decoding methods. All baselines are reproduced under the same settings. \textbf{Bold} indicates the best performance, and \underline{underlined}  indicates the second-best performance.}
   \label{tab:results}
   \vspace{-2mm}
\end{table*}

\begin{figure*}[t]
    \centering
    \includegraphics[width=\linewidth]{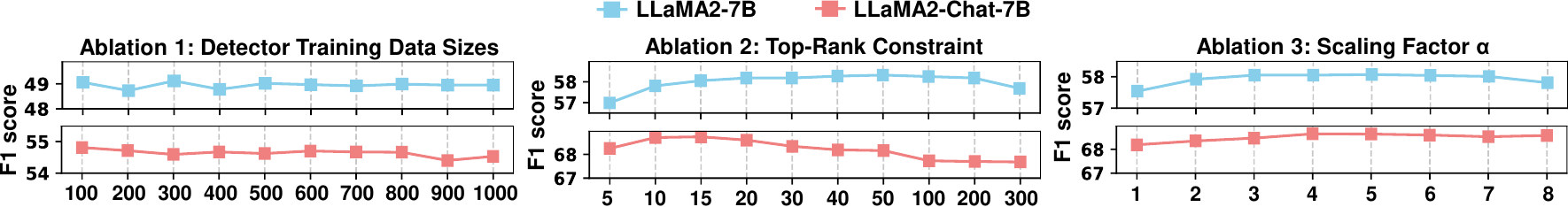}
    \caption{Ablation Study for DAGCD. \textbf{Left Part}: Ablation 1 Detector Training Data Sizes, \textbf{Center Part}: Ablation 2 Top-Rank Constraint, \textbf{Right Part}: Ablation 3 Scaling Factor $\alpha$.}
    \label{fig:ablation-study}
    \vspace{-5mm}
\end{figure*}

\subsection{Experimental Setup} 
\vspace{-2mm}

\noindent \paragraph{Datasets} 
We evaluate context faithfulness on Open-Book Question-Answering (QA) datasets, where each question is paired with external context containing the correct answer. This setup ensures that only context-grounded answers are considered valid, allowing for the assessment of whether the model generates context-faithful hallucinations based on answer accuracy. Specifically, We conducted experiments on seven open-book QA datasets. For further details, refer to Appendix \ref{app:data details}.

\noindent \paragraph{Implementation Details}
\textbf{We used 100 samples constructed from a single dataset (HotpotQA) as the training set to train the Context Utilization Detector.} The scaling factor \( \alpha \) is set to 2 for pre-trained models and 4 for instruction-tuned models to account for entropy variations. The logistic regression classifier and utilization distribution are computed using the top-10 attention heads, with adjustments restricted to the top-10 ranked tokens (\( U_{\text{top}} \)). Further detailed settings see Appendix \ref{app: Prompt Design and Decoding Strategy}.

\noindent \paragraph{Metrics}
Consistent with prior work\cite{jin-etal-2024-tug-of-war,yuan-etal-2024-COIECD,wang2024adacad}, we use EM and F1 score metrics to evaluate the performance of the models on open-book QA datasets.

The LLMs used in our experiments and the baselines are detailed in Appendix \ref{app: used-LLMs} and \ref{app: baseline setting}.

\subsection{Model Performance Comparison}

Table \ref{tab:results} shows DAGCD’s effectiveness across diverse QA datasets and models. \textbf{We also tested DAGCD on summarization tasks, yielding improvements}. For details see Appendix~\ref{app: summary-res}.

\subsubsection{Dataset-Level Observations}
DAGCD achieves consistent improvements across diverse QA tasks, including multi-hop reasoning, long-form retrieval, and document-level QA.

\noindent \paragraph{HotpotQA, TriviaQA, SearchQA:} 
DAGCD excels in multi-paragraph reasoning and long-form retrieval tasks. It achieves the highest gains on HotpotQA with a \textbf{18.80\%} EM and \textbf{14.81\%} F1 improvement on Mistral-7B. DAGCD also outperforms baselines on TriviaQA and SearchQA, showing significant improvements across models.

\noindent \paragraph{SQuAD, NewsQA, NQ:} 
DAGCD demonstrates robust performance in single-paragraph and document-level tasks. On NQ, it achieves a \textbf{71.46\%} EM and \textbf{39.10\%} F1 improvement on Mistral-7B over greedy decoding, while delivering consistent gains across SQuAD and NewsQA datasets.

\noindent \paragraph{NQ-Swap:} 
In adversarial scenarios simulated by NQ-Swap, DAGCD shows notable improvements, including \textbf{74.52\%} EM and \textbf{51.74\%} F1 gains on Mistral-7B, highlighting its robustness.

\subsubsection{Model-Level Observations}
DAGCD demonstrates broad applicability across different model families, sizes, and tuning variants.

\noindent \paragraph{Model Families:}
DAGCD enhances performance across LLaMA and Mistral families. On Mistral-7B, DAGCD improves EM by \textbf{27.86\%} on Mistral-7B and \textbf{14.33\%} on LLaMA2-7B compared to greedy decoding.

\noindent \paragraph{Model Sizes:}
DAGCD improves performance across models of varying model sizes, achieving an average EM and F1 increase of \textbf{14.33\%} and \textbf{11.93\%} on LLaMA2-7B, and \textbf{10.82\%} and \textbf{10.12\%} on LLaMA2-13B, respectively.

\noindent \paragraph{Instruction-Tuned Models:}
Instruction-tuned models, after fine-tuning, show reduced uncertainty in generation probabilities, limiting our method's improvement margin.  However, DAGCD still surpasses all baselines with the highest performance.

\subsection{Ablation Study}
We conducted three ablation experiments to evaluate the impact of variations in different modules on performance. Specifically, in each ablation experiment, the ablation module is adjusted while the other two modules are kept at their default settings.

\noindent \paragraph{Ablation 1: Detector Training Data Sizes}
We tested the impact of detectors trained on different data sizes on actual inference performance. The results, shown in Figure \ref{fig:ablation-study} (left), demonstrate that our method consistently maintains strong performance across various training data sizes, achieving notable results even with just 100 training samples.

\noindent \paragraph{Ablation 2: Top-Rank Constraint}
We evaluated various top-rank constraints on HotpotQA. Figure \ref{fig:ablation-study} (center), top-rank filtering reduces false positives, with F1 score initially improving as constraints loosen, then declining when overly relaxed.

\noindent \paragraph{Ablation 3: Scaling Factor $\alpha$}
We evaluated the impact of different scaling factor \( \alpha \) on model performance. The results, presented in Figure \ref{fig:ablation-study} (right), indicate that \( \alpha \) determines the adjustment intensity applied to the original generation distribution. For pretrained models, optimal performance is achieved at \( \alpha = 2 \), whereas for Chat models, the best performance is observed at \( \alpha = 4 \).

Additional results and performance variations under different prompt templates see Appendix \ref{app: albation-study-details}.

\vspace{-2mm}
\section{Discussion and Analysis}
\vspace{-2mm}

\subsection{Dynamic Decoding: Real-Time Efficiency Without Post-Generation Correction}

Post-generation correction methods, such as CAD~\cite{shi-etal-2024-CAD} and COIECD~\cite{yuan-etal-2024-COIECD}, improve contextual alignment but rely on multi-step processes, causing significant computational overhead. In contrast, DAGCD incorporates context adjustments during generation, ensuring both efficiency and real-time optimization.
\\
\textbf{Lightweight Context Utilization Detector} 
Using a logistic regression-based Context Utilization Detector, our method enables real-time adjustments with minimal computational cost. This detector is more efficient than resource-heavy methods like integrated gradients or attention head manipulation.
\\
\textbf{Single-Pass Decoding with Real-Time Faithfulness Optimization} 
By integrating the Context Utilization Detector directly into the decoding process, our method removes redundant steps like output comparisons or external consistency checks. During generation, attention-based context utilization signals are leveraged in real time to proactively enhance faithfulness. This single-step strategy ensures that the output aligns with the input context without extra post-processing, while maintaining the theoretical time complexity of greedy decoding.

\subsection{Interpretability Through Attention: Insights into Context Utilization}

By systematically analyzing attention mechanisms, our approach uncovers how retrieved context influences the generation process and provides interpretable insights into the behavior of the model.
\\
\textbf{Feature-Based Attention Analysis}
Using natural cases, such as failure in closed-book settings but success in open-book settings, we isolate attention patterns that are indicative of context utilization. A logistic regression classifier trained on these patterns identifies the relevant attention heads with high accuracy, quantifying their contributions.
\\
\textbf{Transparent Decision-Making} The feature coefficients of the logistic regression model directly map to the importance of specific attention heads. This transparency allows for intuitive interpretation, clarifying which heads are most responsible for leveraging context tokens during generation.

\vspace{-2mm}
\section{Related Work}
\vspace{-2mm}

\subsection{Context Faithfulness Hallucination}

Current solutions to context faithfulness hallucination primarily focus on detection and mitigation. For detection, Lei et al.~\cite{hallucination-detection-1} proposed a post-generation editing strategy using natural language inference to classify and revise hallucinated segments. Choi et al.~\cite{hallucination-detection-2} introduced Knowledge-Constrained Decoding, detecting hallucinations during generation and reweighting token distributions to guide output. Chuang et al.~\cite{chuang-etal-2024-lookbacklens} proposed Lookback Lens Guided Decoding, selecting the most faithful output among candidates to improve consistency.

For mitigation, CAD~\cite{shi-etal-2024-CAD} compares outputs with and without concatenated context to enhance contextual adherence, while COIECD~\cite{yuan-etal-2024-COIECD} improves upon CAD by incorporating entropy-based constraints to balance context usage. Wang et al.~\cite{wang2024adacad} further introduced ADACAD, which dynamically adjusts token-level adherence using divergence between contextual and non-contextual outputs.

While effective, most existing methods face limitations such as high computational overhead and reliance on multiple decoding passes. In this work, we propose a real-time solution that integrates context utilization signals directly into the decoding process, achieving efficient and faithful generation, without the need for additional processing steps.

\vspace{-2mm}
\subsection{Attention and Interpretability}

The attention mechanism provides valuable insights into how models prioritize different parts of an input sequence~\cite{2019-Attention-Mechanism-1, 2023-attention-hypothesis-1} and has become central to understanding Transformer-based LLMs~\cite{2019-Attention-Mechanism-2, 2021-Attention-Mechanism-3, 2024-attention-explainability-2}. In LLMs, attention heads often perform distinct roles, such as capturing syntactic dependencies or aligning semantic relationships~\cite{2022-induction-heads, 2024-attention-explainability-1, cutting-off-heads-2024}.

Recent studies have also explored the collaborative behavior of attention heads. For instance, the retrieval head framework~\cite{2024-retrieval-head} identifies heads that collectively retrieve relevant tokens, while cutting-off-heads~\cite{cutting-off-heads-2024} highlights critical heads through systematic ablation. Gradient-based methods like IRCAN~\cite{IRCAN} further investigate the contributions of attention scores and neurons to model outputs.

Unlike previous work, which focused on individual heads, our study examines the collaborative patterns of multiple attention heads. By analyzing how attention mechanisms collectively utilize contextual tokens, we provide a holistic view of their role in aligning outputs with user-provided context.

\vspace{-2mm}
\section{Conclusion}
\vspace{-2mm}

In this paper, we mitigate context faithfulness hallucinations in LLMs by proposing Dynamic Attention-Guided Context Decoding, a lightweight framework that integrates attention distributions and entropy-based uncertainty signals to amplify contextually relevant tokens during generation. Our analysis revealed a strong correlation between high uncertainty and hallucinations, and probing experiments validated that attention mechanisms encode signals indicative of contextual utilization, and further demonstrated that this signal is a fundamental mechanism in Transformer-based LLMs. Experiments across multiple open-book QA datasets demonstrated that DAGCD achieves consistent improvements in context faithfulness, robustness, and scalability, providing an effective solution for context-sensitive generation tasks.


\section*{Limitations}

\paragraph{Dependency on Classifier Accuracy and Robustness to Noisy Contexts}  
DAGCD relies on an attention-ratio based classifier to assess the relevance of context tokens during generation. While the classifier demonstrates high accuracy across datasets and models, its performance may degrade in scenarios with extremely long contexts, complex dialogues, or noisy inputs. Misclassifications in these cases could lead to incorrect adjustments, potentially amplifying irrelevant tokens or diminishing the contribution of critical ones. Similarly, the method’s robustness to adversarial or noisy contexts with misleading or irrelevant information remains an open challenge. Enhancing the classifier’s resilience and incorporating mechanisms to filter or downweight adversarial noise could further strengthen DAGCD’s applicability in real-world scenarios.

\paragraph{Scaling Factor Adjustment for Model Characteristics}  
The scaling factor \( \alpha \) introduced in DAGCD needs to be adjusted based on the characteristics of different models. Although our study shows a strong correlation between entropy-guided uncertainty measures and the model's uncertainty during generation, it does not establish a precise quantitative relationship. This limitation necessitates empirical calibration of the scaling factor for each model to ensure effective adjustments. Such calibration ensures that the method compensates for model-specific entropy variations, but it may introduce additional computational overhead during deployment.

\paragraph{Generalization Across Tasks and Domains}  
Our evaluation primarily focuses on QA tasks, leaving the generalization of DAGCD to other tasks, such as summarization or dialogue generation, unexplored. The attention-ratio based classifier, optimized for QA datasets, may require additional fine-tuning or redesign to handle different output structures and task-specific challenges. Extending the method to diverse domains and tasks could further validate its robustness and scalability.

\bibliography{custom}

\clearpage
\appendix

\section{Details of Experiment "2.1 Uncertainty Leads to Unfaithful Answers"}
\label{app:uncertainty-details}

\subsection{Dataset Used for Analysis}
\label{app:sec_analysis_data}
We randomly sampled 6,000 instances from the MrQA training set~\cite{fisch-etal-2019-mrqa}, which consolidates six open-domain question answering datasets. Specifically, we selected 500 correctly answered and 500 incorrectly answered instances from each sub-dataset. An answer was considered correct if it achieved an Exact Match (EM) with the reference answer.

\subsection{Model Used for Analysis, Prompts, and Decoding Method}

We conduct our analysis using LLaMA2-7B \cite{touvron2023llama2} as the target model. The answers are generated using the following prompt:

"\texttt{Given the following information: \{context\} Answer the following question based on the given information with one or a few words: \{question\} Answer:}"

To ensure deterministic outputs, we employ greedy decoding.

\subsection{Computation Process}

For each sample, we calculated two metrics to quantify uncertainty and confidence. First, the \textbf{Normalized Entropy (NE)} measures the dispersion of probabilities across the vocabulary, providing an overall view of the model's uncertainty. And NE is defined as:
\begin{equation}
H_{\text{norm}}(P) = -\frac{\sum_{i=1}^N P_i \log P_i}{\log N},
\label{eq:normalized_entropy}
\end{equation}

where $P$ represents the token-level probability distribution, and $N$ denotes the vocabulary size.

Second, the \textbf{Maximum Softmax Probability (MSP)} represents the likelihood of the most probable token, offering a complementary perspective on the model's prediction confidence. These metrics focus on the initial generated token distribution to analyze how uncertainty affects response faithfulness.

\subsection{Spearman Correlation Analysis}
\label{app: spearman}

To further investigate the relationship between uncertainty and model errors (i.e., the inability to faithfully respond to the input context), we conducted a Spearman correlation analysis. Specifically, we used the normalized entropy of the token level probability distribution to measure the model's token-level uncertainty, and evaluated answer accuracy using the F1 score. We then analyzed the correlation between answer accuracy and uncertainty during answer generation, with results presented in Table \ref{table:spearman}. Our findings reveal a significant negative correlation, which becomes more pronounced after concatenating the context—i.e., higher uncertainty corresponds to lower answer accuracy (in open-book QA tasks, lower accuracy indicates that the model's response deviates from the provided context). This analysis further suggests that \textbf{token-level uncertainty is strongly associated with unfaithful answers.}

\begin{table}[t] 
  \centering
  \belowrulesep=0pt  
  \aboverulesep=0pt  
    \resizebox{0.8\linewidth}{!}{  
    \begin{tabular}{ccc}
    \toprule[1.5pt]
    \multicolumn{1}{c|}{\textbf{Model}} & \textbf{w/o context} & \textbf{w/ context} \\
    \midrule
    \multicolumn{1}{c|}{LLaMA2-7B} & -0.43  & -0.53  \\
    \multicolumn{1}{c|}{LLaMA2-7B-Chat} & -0.27  & -0.33  \\
    \multicolumn{1}{c|}{LLaMA2-13B} & -0.30  & -0.51  \\
    \multicolumn{1}{c|}{LLaMA2-13B-Chat} & -0.26  & -0.32  \\
    \multicolumn{1}{c|}{Mistral-7B} & -0.23  & -0.22  \\
    \multicolumn{1}{c|}{Mistral-7B-Instruct} & -0.30  & -0.33  \\
    \bottomrule[1.5pt]
    \end{tabular}%
}
    \caption{Spearman correlation analysis. We examine the relationship between F1 scores (pred-ans vs. gold-ans) and norm-entropy (generation distribution) under both w/o and w/ context settings (\( p \ll 0.05 \) for all models).}

    \label{table:spearman}
    
\end{table}%

\subsection{Probability Gap Between The Golden Answer Token and The Ranked Top-1 Token}

\begin{figure}
    \centering
    \includegraphics[width=\linewidth]{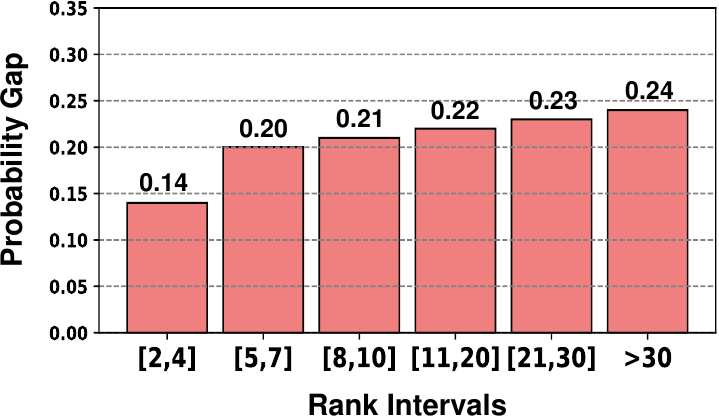}
    \caption{Probability Gap. For incorrect responses, the probability gap between the gold answer token and the highest-probability token.}
    \label{app-fig-probs-gap}
    \vspace{-5mm}
\end{figure}

As shown in Figure \ref{app-fig-probs-gap}, we analyze the wrong answer samples from \ref{app:sec_analysis_data} by calculating the average probability gap between the gold answer token and the highest-probability token. 

\section{Details of "3 Context Utilization Signal in Attention"}
\label{app:section3-details}

\subsection{LR Classifier Training and Evaluation}
\label{app:LR-training-details}
We trained a Logistic Regression (LR) classifier using 5-fold cross-validation. L2 regularization was applied to prevent overfitting. The classifier was evaluated across Transformer-based LLMs, including:
\begin{itemize}
    \item LLaMA2: 7B, 13B, 7B-Chat, 13B-Chat
    \item Mistral: 7B, 7B-Instruct
\end{itemize}

\subsection{Cross Prompt Templates Testing}
\label{app:LR cross Prompt}

To evaluate the robustness of the classifier under different prompts, we reconstructed the attention ratio feature vectors using Prompts 2, 3, and 4. These prompts differ in structure and phrasing but are consistent in task objectives. The classifier, trained using Prompt 1, was then tested on these alternative prompts. (Templates shown in Figure \ref{fig: prompt-template})

The results, shown in Table \ref{table: cross prompt LR}, demonstrate that the classifier maintains an ACC exceeding 97\% and AUC above 0.99 across all prompts. This indicates that the attention ratio signal is prompt-agnostic and generalizes well across different input structures.

\begin{table}[t]
  \centering
  \belowrulesep=0pt  
  \aboverulesep=0pt  
      \resizebox{0.9\linewidth}{!}{  
    \begin{tabular}{c|cc|cc}
    \toprule[1.5pt]
    \multirow{2}{*}{\textbf{Prompt}} & \multicolumn{2}{c|}{\textbf{LLaMA2-7B}} & \multicolumn{2}{c}{\textbf{Mistral-7B}} \\
\cmidrule{2-5}          & \textbf{ACC} & \textbf{AUC} & \textbf{ACC} & \textbf{AUC} \\
    \midrule
    Prompt2 & 0.9797  & 0.9932  & 0.9762  & 0.9902  \\
    Prompt3 & 0.9768  & 0.9926  & 0.9763  & 0.9889  \\
    Prompt4 & 0.9794  & 0.9946  & 0.9771  & 0.9927  \\
    \bottomrule[1.5pt]
    \end{tabular}%
    }
    \caption{Performance testing under different prompts. Training data: attention ratio feature vector under Prompt 1. Test data: attention ratio feature vector under Prompts 2, 3, and 4.}
      \label{table: cross prompt LR}%
      \vspace{-5mm}
\end{table}%

\begin{figure*}[ht]
    \centering
    \includegraphics[width=0.95\linewidth]{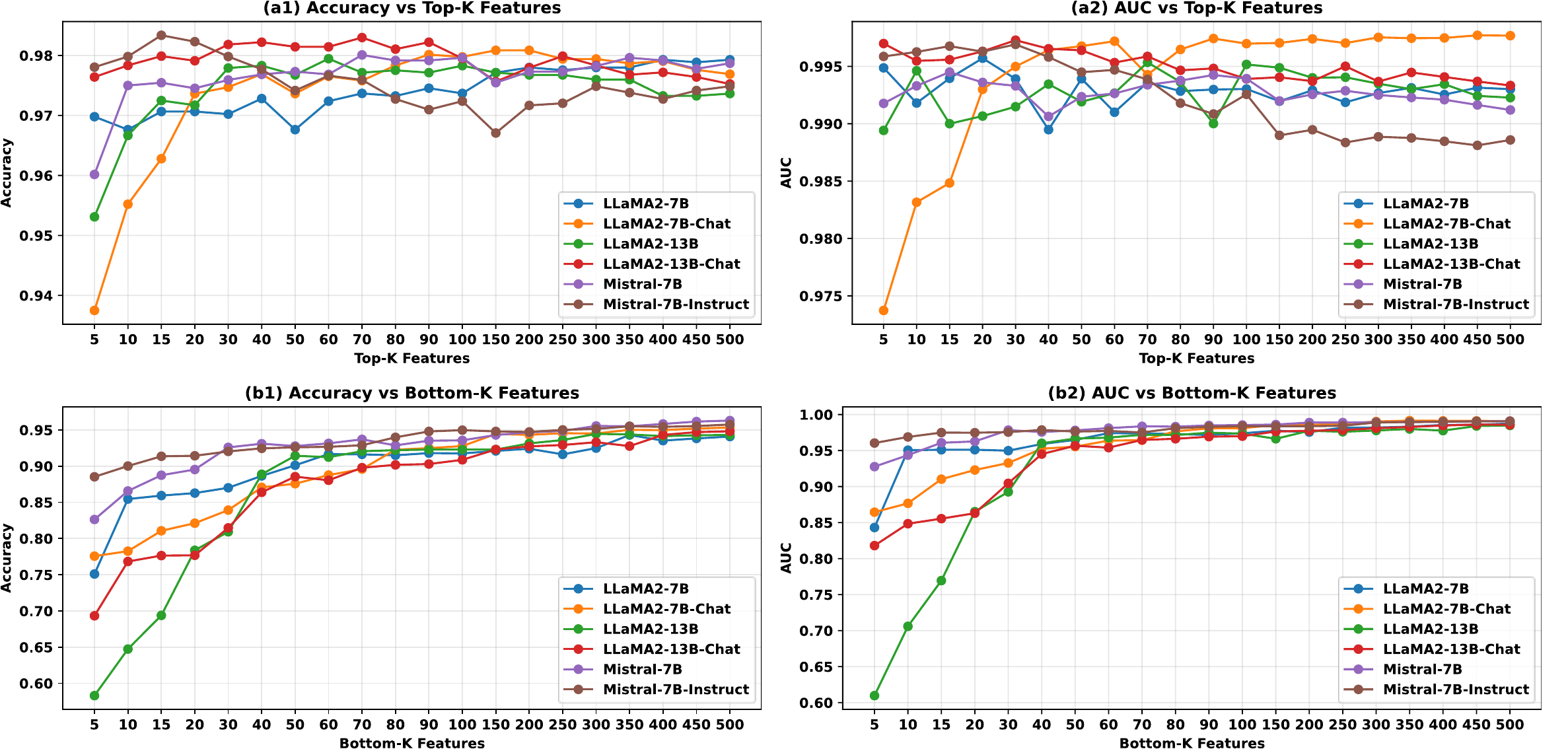}
    \caption{Performance of the LR classifier with Top-\(K\) and Bottom-\(K\) features. Based on the absolute values of feature coefficients, the Top-\(K\) and Bottom-\(K\) features were selected to train an LR classifier with sparse features. The figure shows the ACC and AUC performance of the classifier on the test set for different values of \(K\).}
    \label{app-fig: topk bottomk acc auc}
\end{figure*}

\begin{figure*}[ht]
    \centering
    \includegraphics[width=0.95\linewidth]{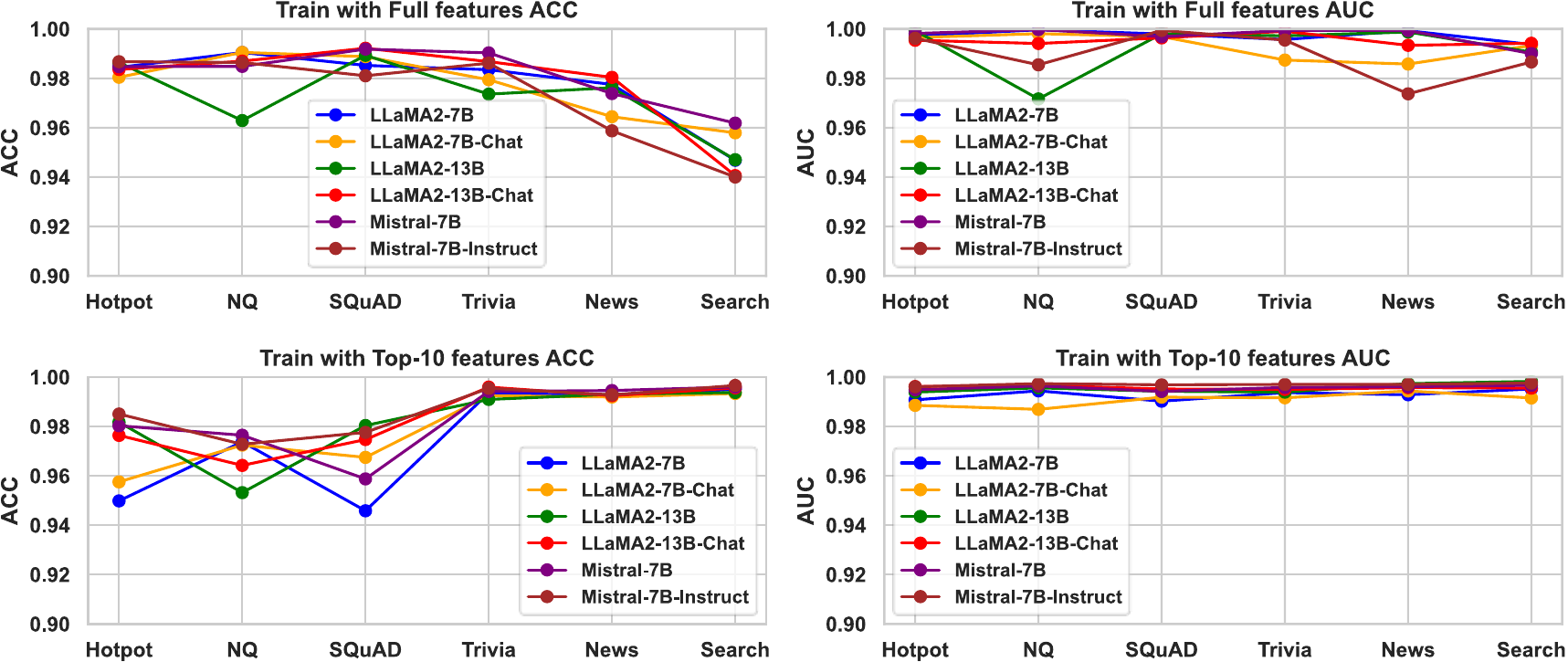}
    \caption{Out-of-domain performance of the LR classifier trained with full features and Top-10 features. Performance variations on out-of-domain data for LR classifiers trained using all features versus the top-10 features.}
    \label{app-fig: topk out-domain}
\end{figure*}

\begin{figure*}[htbp]
    \centering
    \includegraphics[width=0.95\linewidth]{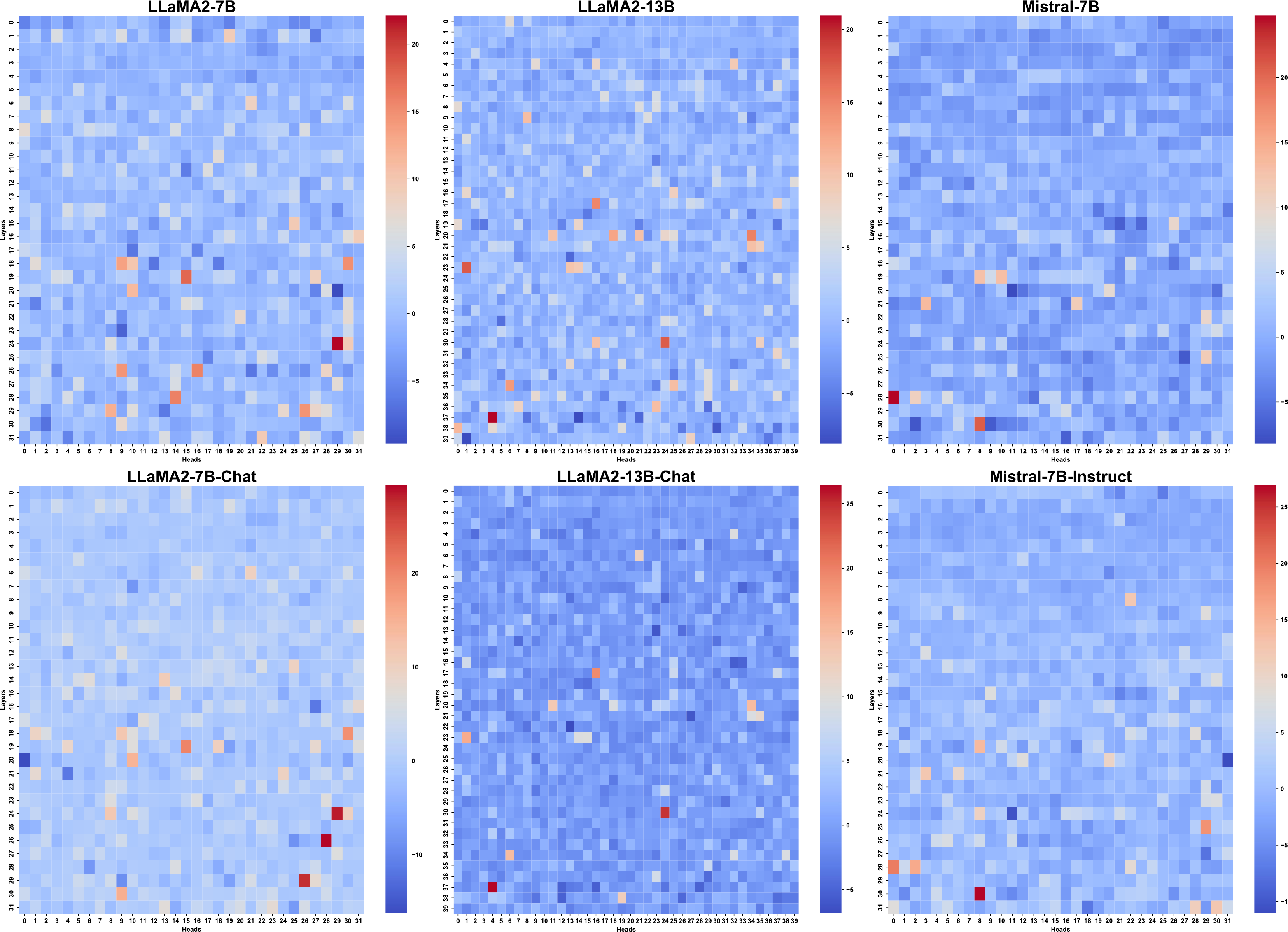}
    \caption{Heatmap of feature coefficients for LR. The heatmap of feature coefficients for LR classifiers trained using attention ratios from different LLMs as features.}
    \label{app-fig: heatmap}
\end{figure*}

\begin{figure*}[htbp]
    \centering
    \includegraphics[width=0.95\linewidth]{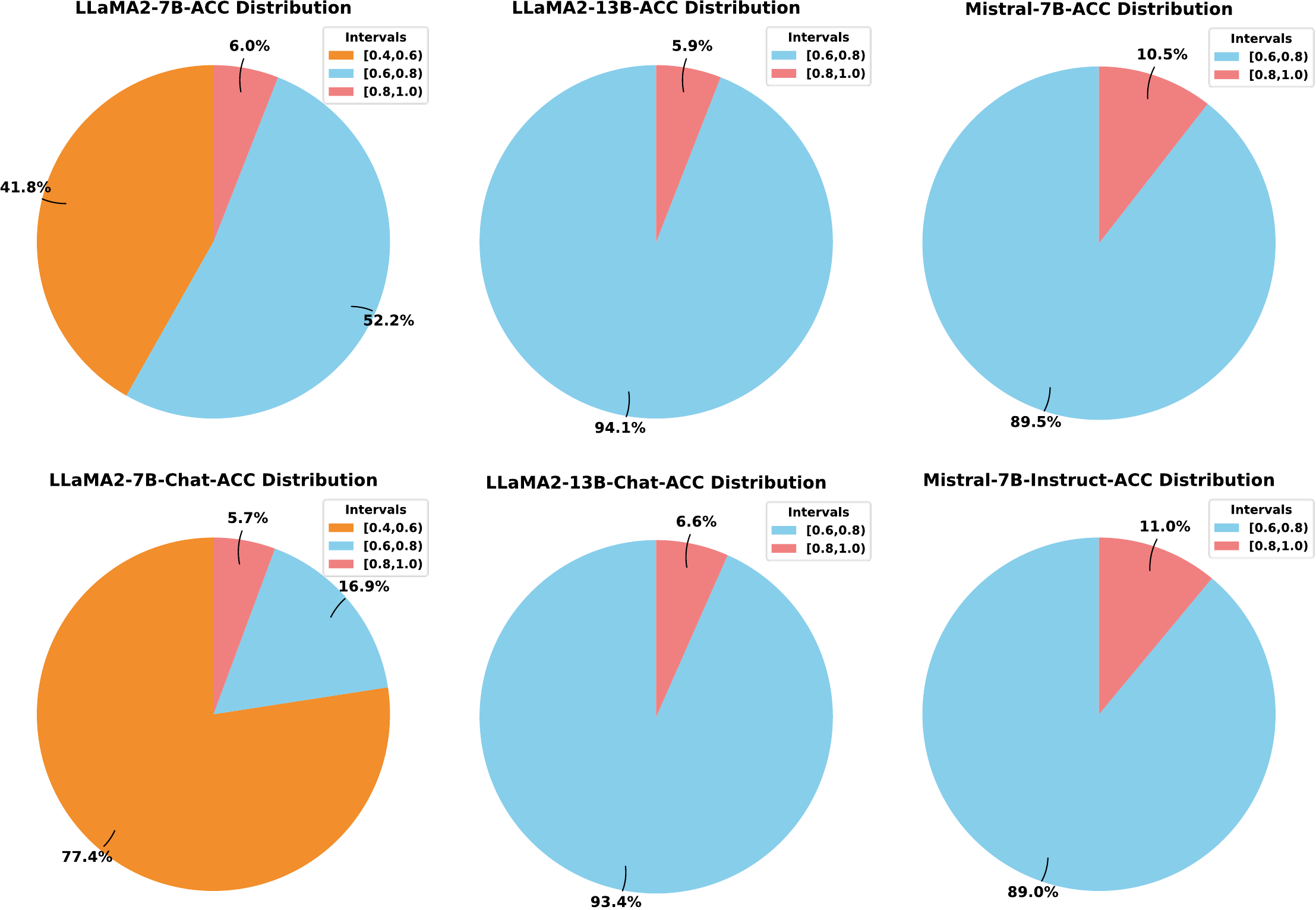}
    \caption{Performance distribution of LR with single features. An LR classifier is trained using the attention ratio from a single head as features. The figure shows the ACC distribution of LR classifiers trained with attention ratios from different heads on the test set.}
    \label{app-fig: single featture}
\end{figure*}

\begin{figure}[htbp]
    \centering
    \includegraphics[width=\linewidth]{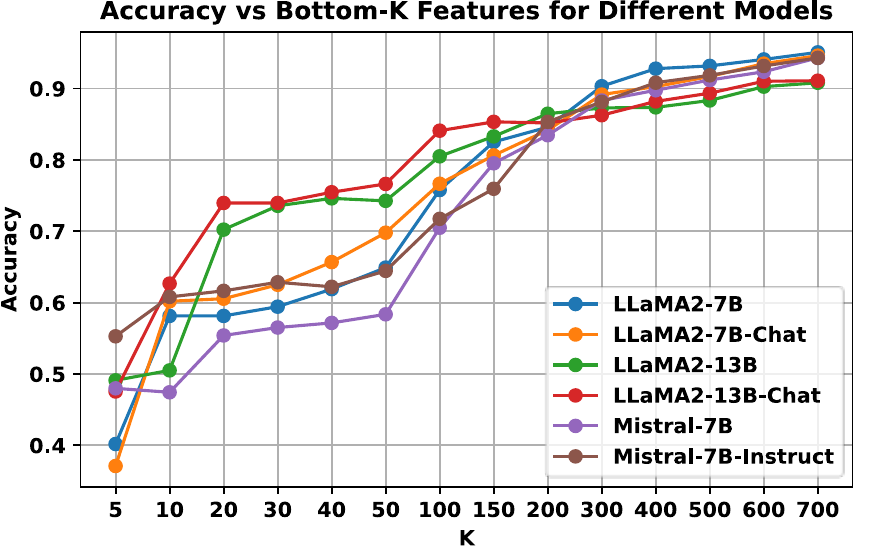}
    \caption{Performance of the LR classifier trained with Bottom-\(K\) features. Based on the ACC of LR classifiers trained using the attention ratio from a single head as features, the Bottom-\(K\) heads with the lowest ACC are selected. The LR classifier is then retrained using the attention ratios of these Bottom-\(K\) heads as features, and its performance on the test set is presented.}
    \label{app-fig: single-bottom-k}
\end{figure}

\section{Analyzing Context Utilization Signal Contribution Across Heads}
\label{app: Characteristics of Different Features}

To further explore the role of attention heads in encoding contextual utilization signals, we conducted a detailed analysis to examine the strength and distribution of signals across heads. This section presents insights into the concentration of strong signals in certain heads, the independent utility of individual heads, and the complementarity of weaker heads.

\subsection{Analysis of the Importance of Different Features}

In analyzing the LR classifier trained with attention values from all attention heads as features, we found that most feature coefficients had small absolute values. This indicates that only a few attention heads are crucial for utilization detection. 

To explore their impact, we selected the top-K and bottom-K features based on the absolute values of their coefficients and trained LR models using these subsets. Figure \ref{app-fig: topk bottomk acc auc}: subplots (a1) and (a2) shows how classification accuracy (ACC) changes with K. Using top-K features, the model achieves over 0.95 accuracy for all LLMs with K=10, matching the performance of using all features. In contrast, models with bottom-K features perform poorly, failing to reach 0.95 even with K=100. The AUC curves in Figure \ref{app-fig: topk bottomk acc auc}: subplots (b1) and (b2) for different K further confirm this. Models with top-K features maintain high accuracy and robustness, while those with bottom-K features show significantly worse performance. These results emphasize that a small number of key attention heads are enough for effective detection, while irrelevant features add little value and may introduce noise. 

We also compared the performance of LR classifiers trained with all features versus only the top-10 features on out-of-domain data. As shown in Figure \ref{app-fig: topk out-domain}, the LR trained with only the top-10 features achieved better ACC and AUC on out-of-domain data.

Based on the ACC and AUC results, we find that \textbf{the LR classifier trained with the Top-10 features achieves good accuracy and robustness while using the minimum number of features}. Therefore, all subsequent inference employs LR classifiers trained with the Top-10 features.


\subsection{Signal Strength and Concentration in Heads}

To identify influential heads, we visualized the coefficients of the trained Logistic Regression (LR) classifier, which were derived from the attention ratio features of all heads. Approximately 5\% of the heads exhibited significantly high coefficients, suggesting that these heads dominate the classification task (Figure \ref{app-fig: heatmap}). Repeating this analysis across 100 random seeds revealed consistent selection of these top heads, indicating their robustness as key signal carriers.

To evaluate the standalone utility of these heads, we trained LR classifiers using the attention ratio from a single head as the feature. The results in Figure \ref{app-fig: single featture} show that About 5\% to 10\% of the heads achieved classification accuracies (ACC) above 0.8, highlighting their ability to independently encode contextual utilization signals. However, the majority of heads performed poorly in isolation, with ACCs below 0.8. This disparity emphasizes the varying degrees of utility across heads, with a small subset contributing disproportionately strong signals.

Additionally, we also observed that on the Mistral model, the vast majority of heads perform well when acting individually. This indicates \textbf{the presence of more high-performing heads in the Mistral model, which may explain why our method achieves greater improvements on Mistral compared to other models}.

\vspace{-2mm}
\subsection{Complementary Contributions of Weaker Heads}
\vspace{-2mm}
Although most attention heads have limited standalone utility, we observe that combining weaker heads into subsets significantly improves classification performance. Figure \ref{app-fig: single-bottom-k} illustrates that we selected the bottom-K features, based on the classification accuracy of individual heads, to train the classifier and analyze the performance gain from combining weaker heads. The results show that when the number of bottom-K heads reaches 500, the classification accuracy stabilizes at approximately 90\%. This finding highlights the complementarity of weaker heads, as their aggregated signals collectively achieve robust token classification.

\paragraph{Summary of Findings} 
Our analysis reveals three key characteristics of attention heads in encoding contextual utilization signals:
\begin{enumerate}
    \item \textbf{Concentration:} A small subset of heads consistently contributes strong independent signals, dominating the classification task.
    \item \textbf{Complementarity:} Weaker heads collectively provide complementary signals, enabling robust classification when aggregated.
\end{enumerate}
These findings highlight the nuanced roles of attention heads in contextual token utilization and provide a foundation for further exploration of their properties and applications.

\section{Experimental Details}
\label{app:exp}

\subsection{Dataset Details}
\label{app:data details}

We conducted experiments on seven open-book question-answering (QA) datasets, representing a variety of QA tasks. These include multi-hop reasoning datasets (HotpotQA \cite{yang-etal-2018-hotpotqa}), long-form retrieval-based QA datasets (TriviaQA \cite{joshi-etal-2017-triviaqa} and SearchQA \cite{dunn2017searchqa}), single-paragraph extraction tasks (SQuAD \cite{rajpurkar-etal-2016-squad} and NewsQA \cite{trischler-etal-2017-newsqa}), and document-level QA datasets (NQ \cite{NQ}). All datasets are formatted in the unified schema provided by the MrQA repository \cite{fisch-etal-2019-mrqa}. Additionally, we used the artificially constructed NQ-swap dataset \cite{longpre-etal-2021-entity(NQ-swap)}, designed to simulate conflicting or ambiguous scenarios by replacing entities.

\subsubsection{Dataset Categories and Statistics}

The datasets used in this study include seven open-book question-answering (QA) datasets, grouped into three categories based on their QA task characteristics: multi-hop reasoning, long-form retrieval-based QA, and single-paragraph extraction tasks. Additionally, an adversarial dataset is included for evaluating the robustness of the proposed method. Detailed descriptions and dataset statistics are provided below.

\noindent \paragraph{Multi-Hop Reasoning (HotpotQA).} HotpotQA \cite{yang-etal-2018-hotpotqa} is a benchmark dataset for multi-hop reasoning, requiring models to synthesize information across multiple paragraphs to generate an answer. This dataset emphasizes complex reasoning over distributed evidence, making it a critical benchmark for evaluating context utilization.

\noindent \paragraph{Long-Form Retrieval-Based QA (TriviaQA, SearchQA).} TriviaQA \cite{joshi-etal-2017-triviaqa} and SearchQA \cite{dunn2017searchqa} require reasoning over longer contexts, with answers scattered across retrieved documents. These datasets test the model’s ability to focus on relevant content in lengthy contexts and generate precise answers.

\noindent \paragraph{Single-Paragraph Extraction (SQuAD, NewsQA).} SQuAD \cite{rajpurkar-etal-2016-squad}, and NewsQA \cite{trischler-etal-2017-newsqa} are standard extractive QA datasets where the answer is typically located within a single paragraph. These datasets are widely used for evaluating the span-extraction capabilities of QA systems.

\noindent \paragraph{Document-Level QA (NQ).} NQ \cite{NQ} is a document-level open-domain question answering dataset driven by real user queries. It requires systems to extract long answers from entire Wikipedia documents and generate specific short answers, evaluating document-level information retrieval and natural language understanding capabilities.

\noindent \paragraph{Simulated Conflict Scenarios (NQ-Swap).} NQ-Swap \cite{longpre-etal-2021-entity(NQ-swap)} is an artificially constructed dataset that introduces adversarial entity swaps into NQ to create ambiguous or conflicting contexts. It evaluates the model's ability to resolve conflicts and faithfully utilize context.

\subsubsection{Dataset Sources and Formats}
All datasets are standardized in the unified schema provided by the MrQA repository (Huggingface ID: mrqa-workshop/mrqa), except for NQ-Swap, which is sourced from a separate repository (Huggingface ID: pminervini/NQ-Swap). The datasets used for training the logistic regression model (§ \ref{sec2}) and attention analysis (§ \ref{sec:attention_signals}) are drawn from the training sets of the MrQA repository. Model performance evaluation is conducted using the validation sets from the same repository. All datasets have been preprocessed to ensure compatibility with our experimental framework.

\subsubsection{Dataset Statistics}
\label{app: dataset-statistic}
Table~\ref{app-tab:data size} presents the size of the datasets used in this study to evaluate model performance.

\begin{table}[t]
  \centering
  \belowrulesep=0pt  
  \aboverulesep=0pt  
    \resizebox{\linewidth}{!}{  
    \begin{tabular}{c|c}
    \toprule
    Dataset & Number of Samples \\
    \midrule
    HotpotQA (Multi-Hop) & 5904 \\
    TriviaQA (Long-Form Retrieval) & 7785 \\
    SearchQA (Long-Form Retrieval) & 16980 \\
    SQuAD (Single-Paragraph) & 10507 \\
    NewsQA (Single-Paragraph) & 4212 \\
    NQ (Document-Level) & 12836 \\
    NQ-Swap (Simulated Conflicts) & 4746 \\
    \bottomrule
    \end{tabular}%
    }
  \caption{Dataset statistics. A summary of the dataset sizes used for evaluation across different datasets.}
  \label{app-tab:data size}%
\end{table}%

\subsection{Implementation Details}
\label{app: Prompt Design and Decoding Strategy}

At each decoding step, DAGCD determines whether utilized tokens are detected by the Context Utilization Detector. If detected, their probabilities are amplified; otherwise, or if a termination condition is met (e.g., the top-1 token is ``\textbackslash n''), probabilities remain unchanged. All experiments utilized a unified prompt template (Prompt 1, as shown in Figure \ref{app: Prompt Template}) to ensure consistency across methods. The prompt format is detailed in Appendix~\ref{app: Prompt Template}. For decoding, greedy decoding was employed to produce deterministic outputs and facilitate direct comparisons across methods. All models run on NVIDIA A100 GPUs.

\subsection{Model Details}
\label{app: used-LLMs}
The LLMs used in this work, along with its HuggingFace ID, is as follows:
\begin{itemize}
    \item LLaMA2-7B: meta-llama/Llama-2-7b-hf
    \item LLaMA2-7B-Chat: meta-llama/Llama-2-7b-chat-hf
    \item LLaMA2-13B: meta-llama/Llama-2-13b-hf
    \item LLaMA2-13B-Chat: meta-llama/Llama-2-13b-chat-hf
    \item Mistral-7B: mistralai/Mistral-7B-v0.1
    \item Mistral-7B-Instruct: mistralai/Mistral-7B-Instruct-v0.1
\end{itemize}

\subsection{Baseline Configurations}
\label{app: baseline setting}

We compare the proposed method DAGCD with three decoding strategies: Greedy Decoding, CAD~\cite{shi-etal-2024-CAD}, and COIECD~\cite{yuan-etal-2024-COIECD}. CAD and COIECD are specifically designed to mitigate context faithfulness hallucination. We implemented the baseline methods with their recommended hyperparameter settings for fair comparisons:
\begin{itemize}[leftmargin=*]
    \item \textbf{CAD}~\cite{shi-etal-2024-CAD}: The contrastive adjustment factor \( \alpha \) was set to 1.
    \item \textbf{COIECD}~\cite{yuan-etal-2024-COIECD}: The entropy regularization parameter \( \lambda \) was set to 0.25, and the contrastive adjustment factor \( \alpha \) was set to 1.
\end{itemize}

\begin{table*}[t]
  \centering
    \belowrulesep=0pt  
  \aboverulesep=0pt  
    \resizebox{0.8\linewidth}{!}{  
    \begin{tabular}{c|c|ccccc}
    \toprule
    \textbf{Model} & \textbf{Decoding} & \textbf{ROUGE-L} & \textbf{factKB} & \textbf{BERT-P} & \textbf{BERT-R} & \textbf{BERT-F1} \\
    \midrule
    \multirow{4}[2]{*}{\textbf{LLaMA2-7B}} & Greedy & 0.2081  & \textbf{0.9932 } & 0.9000  & 0.7997  & 0.8465  \\
          & CAD   & \textbf{0.2361 } & 0.9786  & 0.9054  & \underline{0.8016}  & \underline{0.8514}  \\
          & COIECD & 0.2089  & 0.9845  & \underline{0.9152}  & 0.8014  & 0.8543  \\
          & \textbf{OURs} & \underline{0.2134}  & \underline{0.9856}  & \textbf{0.9210 } & \textbf{0.8026 } & \textbf{0.8576 } \\
    \midrule
    \multirow{4}[2]{*}{\textbf{LLaMA2-7B-Chat}} & Greedy & 0.2368  & \underline{0.9846}  & \underline{0.9056}  & \underline{0.8035}  & \underline{0.8515}  \\
          & CAD   & 0.2082  & 0.9417  & 0.9001  & 0.7997  & 0.8466  \\
          & COIECD & \underline{0.2371}  & 0.9807  & 0.9055  & 0.8034  & 0.8513  \\
          & \textbf{OURs} & \textbf{0.2426 } & \textbf{0.9866 } & \textbf{0.9104 } & \textbf{0.8036 } & \textbf{0.8536 } \\
    \bottomrule
    \end{tabular}%
    }
    \caption{Comparison of evaluation results on CNN/DailyMail. The table compares the evaluation results between greedy decoding and our proposed method on the CNN/DailyMail dataset. \textbf{Bold} denotes the best performance, while \underline{underlined} indicates the second-best performance.}
  \label{app-tab: summary}%
\end{table*}%

\subsection{Results on Summarization Tasks}
\label{app: summary-res}

To validate the performance of our approach on long-form answer generation tasks, we conducted experimental evaluations on the CNN\_DM \cite{pointer-networks/CNN_DM} summarization dataset (we randomly sampled 500 instances from the dataset for evaluation). Similar to prior work \cite{shi-etal-2024-CAD}, we adopted ROUGE-L \cite{rouge}, factKB \cite{factkb}, and BERTScore \cite{bertscore} as comprehensive evaluation metrics to assess both the accuracy and factual consistency of the generated content. The experimental results, as shown in Table~\ref{app-tab: summary}, demonstrate that our method achieves significant improvements on both the pretrained and chat versions of LLaMA2.

\section{Detailed Results of "5.3 Ablation Study"}
\label{app: albation-study-details}

\subsection{Additional Results of "Ablation 1: Detector Training Data Sizes"}

\begin{figure}[ht]
    \centering
    \includegraphics[width=\linewidth]{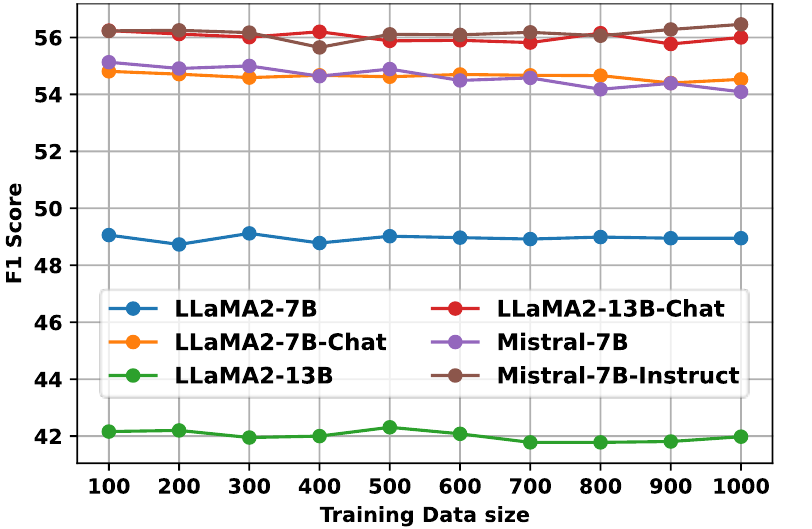}
    \caption{Detector Training Data Size Validation. The variation in inference performance across different models when using detectors trained on varying amounts of data.}
    \label{app: fig-ablation-data-size}
\end{figure}

Table \ref{app: fig-ablation-data-size} shows the performance variations of all LLMs used in this study when trained with detectors on different amounts of data. As observed, our method maintains consistent performance across various data sizes for all LLMs.

\subsection{Additional Results of "Ablation 3: Scaling Factor \texorpdfstring{$\alpha$}{alpha}"}

\begin{figure}[ht]
    \centering
    \includegraphics[width=\linewidth]{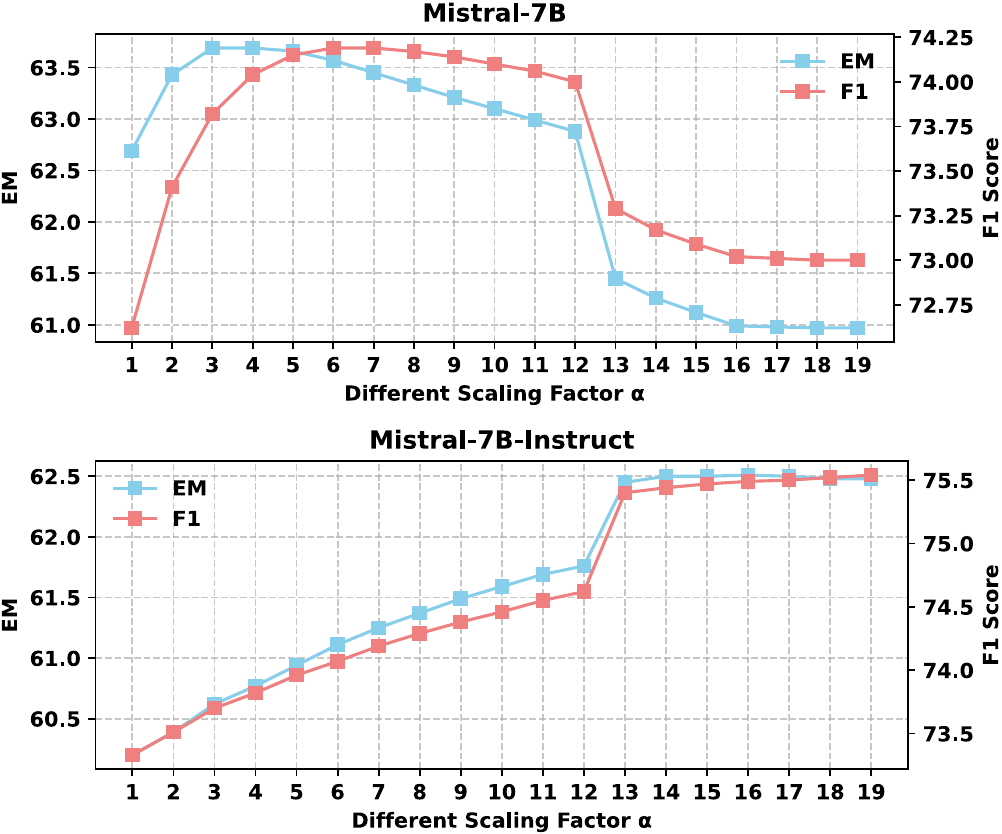}
    \caption{The performance on HotpotQA for DAGCD under different scaling factors.}
    \label{app-fig:ablation Different-alpha}
\end{figure}

We additionally evaluated the performance variations of our method on Mistral-7B and Mistral-7B-Instruct under different scaling factors \(\alpha\). Figure \ref{app-fig:ablation Different-alpha} illustrates the performance changes on the HotpotQA dataset. For Mistral-7B, the optimal performance is achieved at \(\alpha=5\). In contrast, for Mistral-7B-Instruct, the performance only stabilizes after \(\alpha=13\). This indicates that different models may require different optimal scaling factors for the best performance.

\subsection{Impact of Different Prompts}

\begin{figure*}[ht]
    \centering
    \includegraphics[width=\linewidth]{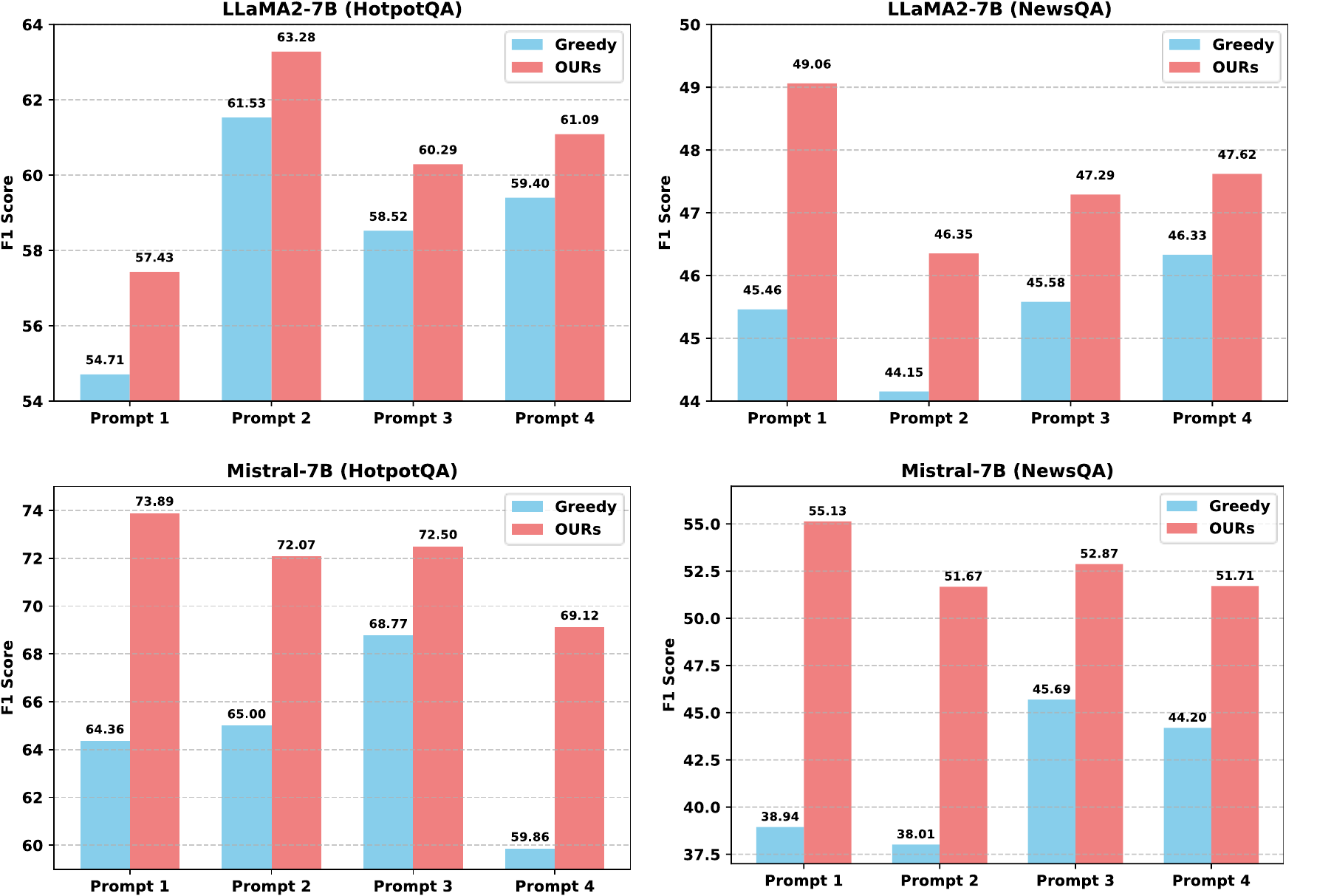}
    \caption{Performance variations across different prompt templates. The figure shows F1 score variations on the HotpotQA and NewsQA datasets for Greedy Decoding and OURs (DAGCD) under different prompt templates.}
    \label{app-fig:ablation Different-prompt}
\end{figure*}

To assess robustness to prompt variations, we tested multiple prompts from prior studies~\cite{zhou-etal-2023-context-faithful, yuan-etal-2024-COIECD, wang2024adacad} (templates in Figure \ref{fig: prompt-template}). Figure \ref{app-fig:ablation Different-prompt} illustrates the variations in F1 scores for LLaMA2-7B and Mistral-7B on the HotpotQA and NewsQA datasets under different prompt templates. The results show that DAGCD consistently outperforms baselines across all tested prompts, demonstrating its adaptability to diverse input formats and reliability across QA tasks.

\section{Prompt Templates}
\label{app: Prompt Template}
\begin{figure*}[!ht]
    \centering
    \includegraphics[width=\linewidth]{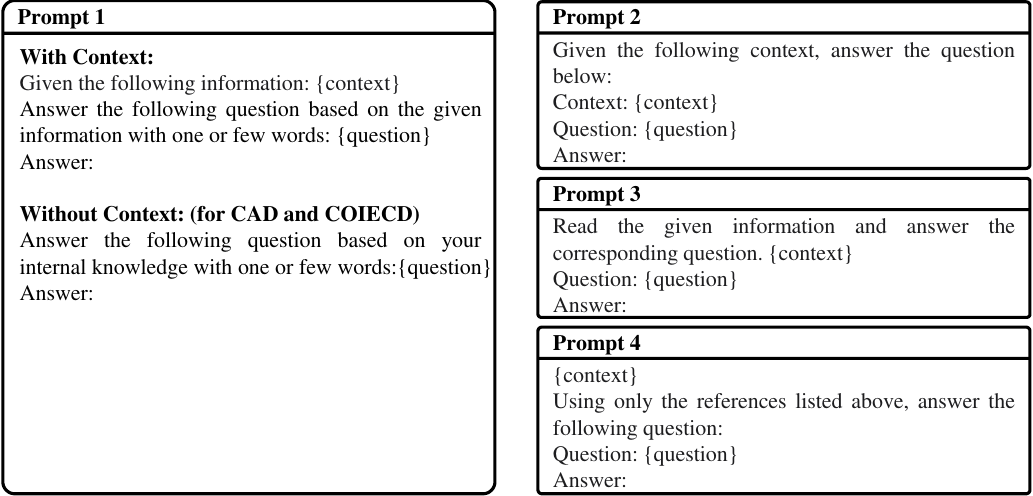}
    \caption{Prompt templates used in this paper.}
    \label{fig: prompt-template}
\end{figure*}

Figure \ref{fig: prompt-template} shows the prompt templates used in this paper.

\end{document}